\documentclass{article}

\usepackage{arxiv}

\usepackage[utf8]{inputenc} 
\usepackage[T1]{fontenc}    
\usepackage{hyperref}       
\usepackage{url}            
\usepackage{booktabs}       
\usepackage{amsfonts}       
\usepackage{nicefrac}       
\usepackage{microtype}      
\usepackage{lipsum}
\usepackage{graphicx}
\graphicspath{ {./images/} }
\usepackage{amssymb}
\usepackage{amsthm}

\usepackage{lineno}
\usepackage{subfigure}
\usepackage{xcolor}
\usepackage{amsmath}

\usepackage{fancyhdr}
\usepackage{lastpage}
\newtheorem{theorem}{Theorem}

\title{Deep Ritz method with Fourier feature mapping: A deep learning approach for solving variational models of microstructure}

\author{
	Ensela Mema \\
	Kean University\\
	Union, NJ 07083 \\
	\texttt{emema@kean.edu} \\
	\And
	Ting Wang\\
	Booz Allen Hamilton Inc.\\
	McLean, VA 22102\\
	\texttt{wang\_ting@bah.edu} \\
	\And
	Jaroslaw Knap \\
	DEVCOM Army Research Laboratory\\
	Aberdeen Proving Ground, MD 21005 \\
	\texttt{jaroslaw.knap.civ@army.mil} \\
}
\begin{document}
	\maketitle
	\begin{abstract}
		This paper presents a novel approach that combines the Deep Ritz Method (DRM) with Fourier feature mapping to solve minimization problems comprised of multi-well, non-convex energy potentials. These problems present computational challenges as they lack a global minimum.  Through an investigation of three benchmark problems in both 1D and 2D, we observe that DRM suffers from spectral bias pathology, limiting its ability to learn solutions with high frequencies. To overcome this limitation, we modify the method by introducing Fourier feature mapping. This modification involves applying a Fourier mapping to the input layer before it passes through the hidden and output layers. Our results demonstrate that Fourier feature mapping enables DRM to generate high-frequency, multiscale solutions for the benchmark problems in both 1D and 2D, offering a promising advancement in tackling complex non-convex energy minimization problems.
	\end{abstract}


\section{Introduction}

Materials undergoing martensitic phase transformations constitute a
technologically important class of
materials~\cite{bhattacharya2003microstructure}. These materials
include steels, shape-memory alloys, solidified gases and polymers, to
name a few. A feature common to all of these materials is
microstructure in the form of elaborate three-dimensional patterns at
the scale ranging from nanometers to centimeters. Mathematically,
microstructure induced by martensitic phase transformations is
characterized as minimizers of a total energy functional.  The
fundamental difficulty in seeking such minimizers lies, however, in
non-convexity of the total energy
functional~\cite{bhattacharya2003microstructure,dacorogna2007direct}.

Numerical treatment of non-convex minimization problems is fraught
with challenges. Standard finite elements usually require very fine
meshes to resolve meaningful scales associated with microstructure. In
addition, specially crafted meshes are frequently needed as finite
element solutions tend to be strongly mesh dependent and adaptive mesh
refinement may not always perform
satisfactorily~\cite{luskin1996computation,Carstensen_2005}. The
strong mesh dependence of solutions may be somewhat alleviated by
recourse to specialized finite-element techniques, such as
discontinuous finite
elements~\cite{gobbert1999discontinuous}. Alternatively, the
non-convex energy functional can be regularized through
convexification~\cite{carstensen2001numerical}. Solutions of
convexified minimization problems can be then efficiently carried out
by standard finite elements~\cite{bartels2004effective}. In practice,
however, convexified energy functionals may not be readily available
explicitly and their numerical approximations are generally costly to
obtain~\cite{carstensen1997numerical}. While minimizers of the
convexified energy functional are much easier to get, they may miss
some important physical features of the original (non-convex) minimization
problem. Finally, one may employ Young measures to turn the non-convex
minimization problem into a convex minimization
problem~\cite{nicolaides1993computation,aranda2001numerical}. This
approach offers numerous benefits, chiefly among them that the energy
functional does not need to be altered. Yet, additional numerical
algorithms are required, increasing considerably the overall
computational
cost~\cite{carstensen2000numerical,bartels2004effective}.

Recent advancements in deep neural networks (DNNs) have raised hopes
that DNNs may be capable of generating solutions to non-convex
minimization problems. Specifically, the universal approximation
theory~\cite{hornik1990universal, hornik1991approximation} has enabled
DNN-based numerical methods for PDEs to parameterize the solution
using a DNN and learn it using the method of stochastic gradient
descent. The approach learns the solution by minimizing a loss
function induced by the physics constraints, often referred to as the
physics informed approach.  Depending on how the loss function is
constructed, DNN-based methods can be roughly classified into three
categories: 1) the physics informed neural network
(PINN)~\cite{raissi2019physics, sirignano2018dgm}; 2) deep Ritz
methods (DRM)~\cite{yu2018deep} and 3) deep backward stochastic
differential equation (BSDE)~\cite{han2017deep}.  PINN minimizes the
residual of the PDE evaluated at a set of randomly sampled collocation
points. In comparison, DRM utilizes the variational structure of
elliptic PDEs to minimize the energy functional.  Finally, deep BSDE
explores the probabilistic connection between parabolic PDE and BSDE
in order to reformulate the problem as a reinforcement learning task.
The key advantage of the DNN-based methods over the conventional ones
lies in the fact that they replace the deterministic mesh by Monte
Carlo sampling and hence, in principle, lead to dimension independent
convergence rates~\cite{grohs2018proof}. Despite being a promising
direction, training of DNN-based methods can be extremely challenging
due to, e.g., the choice of the learning rate, the multi-scale nature
of the problem under consideration, etc. Indeed, it has been widely
observed that DNNs are biased to learn low frequency features of the
solution, making them fail to learn solutions that exhibit
high-frequency and multi-scale, an essential feature in non-convex
minimization in the context of microstructure evolution. This
phenomenon is known as the spectral bias pathology for deep
learning~\cite{rahaman2019spectral, wang2021eigenvector}.

In this work, we focus on the following minimization problem:
\begin{align}
	\min_{u\in \mathcal{U}} \ I(u) \qquad {\rm where } \qquad I(u) = \int_{D} W({\bf x},u({\bf x}), \nabla{u}({\bf x})) \ d{\bf x},\label{eqn:variationalproblem}
\end{align}
where $D \subset \mathbb{R}^d$ is a bounded open set with a Lipschitz
boundary $\partial D$,
$W:\mathbb{R}^d \times \mathbb{R}^N \times \mathbb{R}^{dN} \to
\mathbb{R}$ is the Lagrangian and $u : \bar{D} \to
\mathbb{R}^N$. $\bar{D}$ denotes the closure of $D$.  Here,
$\mathcal{U}$ is a space of admissible functions, e.g., the Sobolev
space $H_0^1(D)$ when the zero boundary condition is imposed. The
energy density $W$ is generally assumed to be non-convex in
$\nabla{u}$.  To solve the above minimization, one seeks minimizers
$u({\bf x})$ of the functional $I(u)$ over the prescribed domain $D$,
subject to boundary condition constraints (set to $u({\bf x}) = 0$ on
$\partial D$). The reader is referred to any standard texts on
variational calculus, for example ~\cite{dacorogna2007direct}, for the
properties of the minimization problem~(\ref{eqn:variationalproblem}).

Since DRM works by minimizing an energy functional, it is natural to
seek solutions of the minimization
problem~\eqref{eqn:variationalproblem} by means of DRM.  A
straightforward application of DRM to non-convex minimization problems
in 1D and 2D has been carried out by Chen et
al. in~\cite{Chen&Rosakis2023}. They demonstrate that DRM is capable
of capturing the complexities of local or global minimizers of
non-convex variational problems, if one applies an ad hoc activation
function.  Additionally, they suggest that the depth of the DNN plays
a role analogous to the mesh size in FEM so one can capture
high-frequency solutions (with more twin bands) if one increases the
depth of DNN. It is important to note that although DRM is capable of
solving non-convex minimization problems, a naive application of the
method fails to consistently generate high-frequency solutions due to
the fact that DNN algorithms, including DRM, are biased to learn the
low frequency features of the solutions.

In our work, we address the shortcomings of DRM by applying Fourier
feature mapping as outlined in~\cite{FourierFeatures2020} and show
that DRM in conjunction with Fourier feature mapping (DRM\&FM) can
consistently generate high-frequency multiscale solutions for
non-convex minimization problems independently of the depth of the
DNN. The main contributions of our work can be summarized as follows:
\begin{itemize}
	\item We apply neural tangent kernel (NTK) theory to show that, similar to PINN, DRM also
	suffers from spectral bias pathology. That is, the learning rates
	along different directions are determined by the corresponding
	eigenvalues of the NTK. To alleviate this issue, we utilize the
	Fourier feature mapping to map the input into an appropriate
	submanifold. Based on the recent theoretical results on
	NTK~\cite{geifman2020similarity, chen2020deep}, we show (at least in the $1$D
	case) that the Fourier feature mapping leads to a quadratic decay
	NTK eigenspectrum which could be advantageous when multiscale
	problems are considered.
	
	\item We numerically illustrate that DRM alone cannot consistently
	generate high-frequency solutions to non-convex minimization
	problems by increasing the depth of DNN. See
	Section~\ref{sec:numerics} for the benchmark problems considered in
	this work and how they differ from the ones considered
	in~\cite{Chen&Rosakis2023}.
	
	\item We apply Fourier feature mapping on DRM and observe that DRM in
	conjunction with Fourier features (DRM\&FM) allow the DNN to learn
	high-frequency solutions to non-convex variational problems
	independently of the depth of the NN.
\end{itemize}
The paper is organized as follows: Section \ref{sec:DRM} outlines how
the DRM can be applied to solve variational problems.  Section
\ref{sec:NTK} uses NTK theory to show that DRM alone suffers from
spectral bias pathology and how Fourier feature mapping enables the
DRM to learn solutions whose NTK has a fast decaying
eigenspectrum. Section~\ref{sec:Numerics} presents our numerical
results in $1D$ and $2D$ and Section \ref{sec:Concl} discusses our
conclusions.

\section{Deep Ritz Algorithm}
\label{sec:DRM}

\begin{figure}[!ht]
	\centering
	\includegraphics[width=0.75\textwidth]{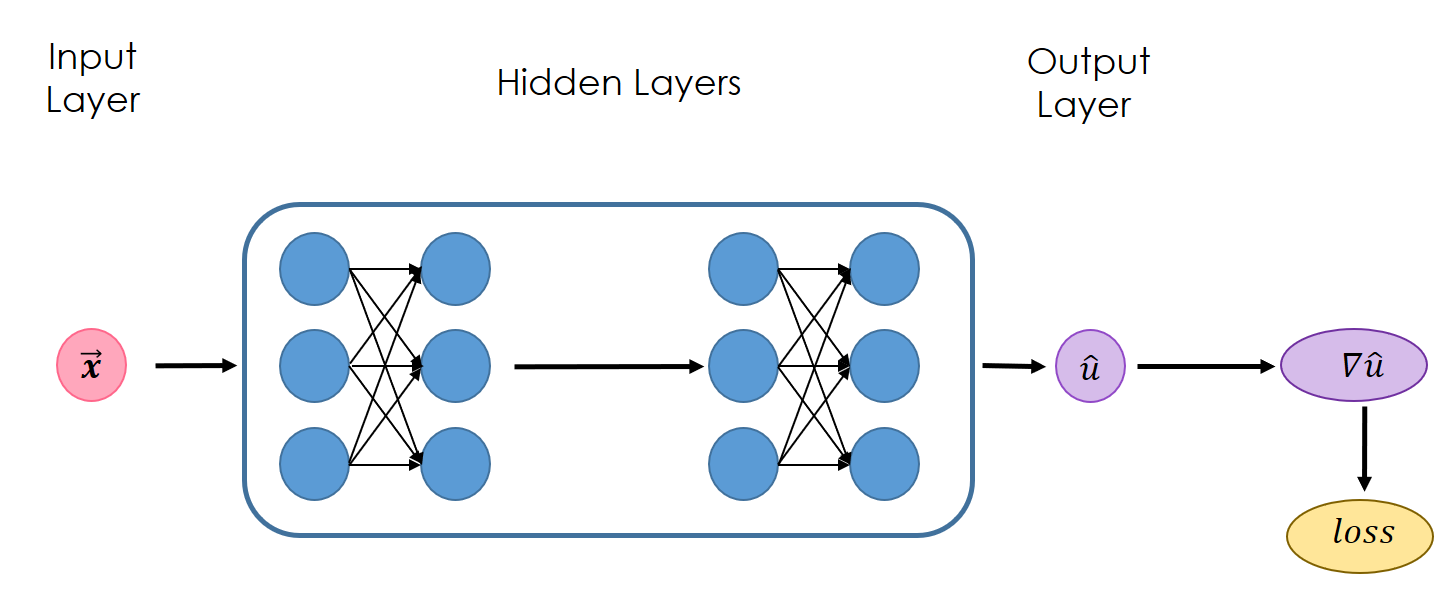}
	\caption{Structure of Neural Network in Deep Ritz Method.}
	\label{fig:NN_structure}
\end{figure}

DRM solves the variational problem in~\eqref{eqn:variationalproblem}
by using DNN to construct an approximation $\hat{u}({\bf x})$ that
minimizes the functional $I(\hat{u})$ over the prescribed domain $D$.
More specifically, a DNN of depth $n$ approximates the solution
through a series of transformations by
\begin{align}
	\hat{u}({\bf x};\theta) = L^{[n]}\circ L^{[n-1]}\circ \cdots \circ L^{[1]}({\bf x}), \label{eqn:NNsolU}
\end{align}
with
\begin{equation*}
	\begin{split}   
		L^{[1]}({\bf x}) &= \sigma(A^{[1]} {\bf x} + b^{[1]})\\
		L^{[i]}({\bf x}) &= \sigma(A^{[i]} L^{[i-1]}({\bf x}) + b^{[i]}), \qquad i = 2, \ldots, n-1\\
		L^{[n]}({\bf x}) &= A^{[n]} L^{[n-1]}({\bf x}) + b^{[n]}
	\end{split}
\end{equation*}
where $A^{[i]}$ and $b^{[i]}$ are the weight matrix and the bias vector of layer $i$, respectively, and $\sigma$ is a nonlinear activation function (see Figure~\ref{fig:NN_structure} for sketch).
Substituting~\eqref{eqn:NNsolU} in the variational problem~\eqref{eqn:variationalproblem} leads to the following finite dimensional optimization problem: 
\begin{align}
	\min_{ \theta \in \mathbb{R}^{N_\theta}} I(\hat{u})\qquad {\rm where} \qquad I(\hat{u}) = \int_{D} W({\bf x},\hat{u}({\bf x};\theta), \nabla \hat{u}({\bf x};\theta) ) d{\bf x},
	\label{eqn:NNoptproblem}
\end{align}
where $\theta = (A^{[1]}, b^{[1]}, \ldots, A^{[n]}, b^{[n]})$ are parameters of the DNN.
To account for the boundary condition, we follow E et al. in ~\cite{Weinan} and Chen et al. in~\cite{Chen&Rosakis2023} in using penalty approach to numerically enforce the prescribed boundary conditions of the variational problem, which leads to a modified functional: 
\begin{align}
	I(\theta) = \int_{D} W({\bf x},\hat{u}({\bf x};\theta), \nabla \hat{u}({\bf x};\theta)) d{\bf x} + \lambda \int_{\partial D} \hat{u}({\bf x};\theta)^2 ds,
\end{align}
where, with a slight abuse of notation, we have rewritten $I(\theta) \triangleq I(\hat{u}, \nabla \hat{u})$ to indicate that the optimization is with respect to the NN parameters $\theta$.
Note that $\lambda $ serves as a penalty term that increases the value of $I$ if the approximated DNN solution, $\hat{u}({\bf x};\theta)$ deviates from the prescribed values at the boundary. 
To solve the optimization problem by stochastic gradient descent (SGD), it is often convenient to rewrite the above integral in its probabilistic form as 
\begin{equation}\label{eqn:prob-form}
	\begin{split}
		\min_{\theta \in \mathbb{R}^{N_{\theta}}} I(\theta) := \mathbb{E} \left[ W({\bf x}, \hat{u}({\bf x}; \theta), \nabla \hat{u}({\bf x}; \theta))\right] + \lambda \mathbb{E}_b \left[|\hat{u}({\bf x_b};\theta)|^2\right],
	\end{split}
\end{equation}
where $\mathbb{E}$ and $\mathbb{E}_b$ are taken with respect to the uniform distributions
over $D$ and $\partial D$, respectively.
At each gradient descent iteration, we use Adam optimizer~\cite{kingma2014adam} to update the DNN parameters $\theta$ by evaluating the stochastic gradient of $I$ at a mini-batch of samples over $D$ and $\partial D$.

\section{NTK analysis for Deep-Ritz and the Fourier feature}\label{sec:NTK}

\subsection{The spectral bias pathology for DRM}
In practice, a naive application of DRM often fails to achieve desirable results.   
In this section, we derive the NTK theory for DRM and show that, similar to PINN, DRM also suffers the pathology of spectral bias of neural networks~\cite{jacot2018neural, rahaman2019spectral, du2018gradient} and hence additional tricks and treats have to be applied. 

For ease of presentation, we assume $d = N = 1$ to keep notation uncluttered.
However, we emphasize that the result presented below can be readily generalized to the vectorial setting.
We start by considering the empirical approximation to~\eqref{eqn:prob-form} without the penalty term, i.e., 
\begin{equation}\label{eqn:ensemble-loss}
	\begin{split}
		\min_{\theta \in \mathbb{R}^{N_{\theta}}} I_{\mathcal{X}}(\theta) := \frac{1}{|\mathcal{X}|} \sum_{{\bf x_n} \in \mathcal{X}} W({\bf x_n}, \hat{u}({\bf x_n}; \theta),  \hat{u}^{\prime}({\bf x_n}; \theta)),
	\end{split}
\end{equation}
where $\mathcal{X}$ is the set of collocation points sampled uniformly over $D$, and $|\mathcal{X}|$ denotes the cardinality of the set.
Applying the gradient descent algorithm to $I_{\mathcal{X}}(\theta)$ leads to the discrete time dynamics 
\[
{\theta}_{n+1} = \theta_n - \eta h \nabla_{\theta} I_{\mathcal{X}}(\theta_n), \qquad n = 1, 2, \ldots,
\]
where $\eta>0$ is the learning rate and $h>0$ is a scaling constant. 
Upon taking $h \to 0^+$, we obtain the continuous time dynamics governing the evolution of the parameters $\theta$,
\begin{equation}\label{eqn:theta-trajectory}
	\frac{d\theta(t)}{dt} = -\eta \nabla_{\theta}I_{\mathcal{X}}(\theta(t)),
\end{equation}
where $\theta: [0, \infty) \to \mathbb{R}^{1 \times N_{\theta}}$ is a function of $t$ and $\nabla_{\theta}I_{\mathcal{X}}(\theta(t)) \in \mathbb{R}^{1 \times N_{\theta}}$ is the gradient of $I_{\mathcal{X}}(\theta)$ with respect to $\theta$. 
We first derive the empirical evolution of the loss function $I_{\mathcal{X}}(\theta(t))$ with respect to $t$, i.e.,  
\begin{equation}\label{eqn:NTK-1}
	\frac{d I_{\mathcal{X}}(\theta(t))}{dt} = \left\langle \nabla_{\theta} I_{\mathcal{X}}(\theta(t)), \frac{d\theta(t)}{dt} \right\rangle = -\eta \|\nabla_{\theta} I_{\mathcal{X}}(\theta(t))\|^2.
\end{equation}
By chain rule we have 
\[
\nabla_{\theta} I_{\mathcal{X}}(\theta) = 
\frac{1}{|\mathcal{X}|} \sum_{{\bf x_n} \in \mathcal{X}} \partial_{\hat{u}} W_n(\theta) \nabla_{\theta} \hat{u}_n(\theta) 
+  \partial_{\hat{u}^{\prime}} W_n(\theta) \nabla_{\theta} \hat{u}^{\prime}_n( \theta), 
\] 
where we have denoted $\hat{u}_n(\theta) = \hat{u}({\bf x_n}; \theta)$, $\hat{u}^{\prime}_n( \theta) = \hat{u}^{\prime}({\bf x_n} ;\theta)$ and 
$W_n(\theta) = W({\bf x_n}, \hat{u}({\bf x_n}; \theta), \hat{u}^{\prime}({\bf x_n}; \theta))$ so that 
\[
\nabla_{\theta} \hat{u}_n \in \mathbb{R}^{1 \times N_{\theta}}, \qquad \partial_{\hat{u}} W_n \in \mathbb{R}, \qquad
\nabla_{\theta} \hat{u}_n^{\prime} \in \mathbb{R}^{1 \times N_{\theta}}, \qquad
\partial_{\hat{u}^{\prime}} W_n \in \mathbb{R}.
\]
Denote 
$
U_n(\theta) = [\hat{u}_n(\theta),  \hat{u}_n^{\prime}(\theta)]^{\top}
\in 
\mathbb{R}^{2 \times 1}
$
so that 
\[
\nabla_{\theta} U_n = [\nabla_{\theta}\hat{u}_n,  \nabla_{\theta}\hat{u}_n^{\prime} ]^{\top} \in \mathbb{R}^{2 \times N_{\theta}},
\]
\[
\nabla_U W_n = [\partial_{\hat{u}} W_n, \partial_{\hat{u}^{\prime}} W_n]^{\top}
\in 
\mathbb{R}^{2 \times 1}
\]
and hence
\[
\nabla_{\theta} I_{\mathcal{X}}(\theta) = \frac{1}{|\mathcal{X}|} \sum_{{\bf x_n} \in \mathcal{X}}  [\nabla_U W_n(\theta)]^{\top} \nabla_{\theta} U_n(\theta) \in \mathbb{R}^{1 \times N_{\theta}}.
\]
Then we can further rewrite the evolution equation given by~\eqref{eqn:NTK-1} in the following compact form,
\begin{equation}\label{eqn:loss-trajectory}
	\begin{split}
		\frac{d I_{\mathcal{X}}(\theta(t))}{dt} 
		= -\frac{\eta}{|\mathcal{X}|^2} \sum_{{\bf x_m}, {\bf x_n} \in \mathcal{X}} [\nabla_U W_m(\theta(t))]^{\top} 
		\left\{\nabla_{\theta} U_m(\theta(t)) [\nabla_{\theta} U_n(\theta(t))]^{\top} \right\}
		\nabla_U W_n(\theta(t)).
	\end{split}
\end{equation}
We call the operator/matrix valued function $K: D  \times D  \to \mathbb{R}^{2 \times 2}$ defined by
\[
K({\bf x_m}, {\bf x_n}; \theta) \triangleq \nabla_{\theta} U_m(\theta) [\nabla_{\theta} U_n(\theta)]^{\top},\qquad {\bf x_m}, {\bf x_n} \in D,
\]
the NTK (parameterized at $\theta$) associated to DRM. 
It should be emphasized that, similar to PINN, the NTK kernel $K$ of DRM depends on both the output $\hat{u}$ and its spatial derivative $ \hat{u}^{\prime}$.

The lazy training phenomenon suggests that, when trained with gradient-based optimizers, strongly overparameterized NNs could converge
exponentially fast to the minimum training loss without significantly varying the parameters~\cite{chizat2019lazy}, i.e., $\theta(t) \approx \theta_0$. 
Therefore, to analyze the asymptotic behavior of the differential equation~\eqref{eqn:loss-trajectory}, we linearize the DNN solution $\hat{u}({\bf x}; \theta)$ at its initial value $\theta_0$ via
\[
\hat{u}({\bf x}; \theta) \approx \bar{u}({\bf x}; \theta) \triangleq \hat{u}({\bf x}; \theta_0) + \langle \nabla_{\theta} \hat{u}({\bf x}; \theta_0), \theta - \theta_0 \rangle,
\]
where by definition $\bar{u}({\bf x}; \theta)$ is the linearization of $\hat{u}({\bf x}; \theta)$ at $\theta_0$.
Notice that 
\[
\nabla_{\theta} [\bar{u}({\bf x}; \theta),  \bar{u}^{\prime}({\bf x}; \theta)]^{\top} = 
\nabla_{\theta}[ \hat{u}({\bf x}; \theta_0),  \hat{u}^{\prime}({\bf x}; \theta_0)]^{\top} 
= \nabla_{\theta} U({\bf x}; \theta_0).
\]
Substituting $\hat{u}({\bf x}; \theta)$ by the linearized model $\bar{u}({\bf x}; \theta)$ into~\eqref{eqn:loss-trajectory} and applying the lazy training assumption to the NTK
leads to the linearized loss dynamics
\begin{equation}\label{eqn:loss-trajectory-linearize}
	\begin{split}
		\frac{d \bar{I}_{\mathcal{X}}(\theta(t))}{dt} 
		= -\frac{\eta}{|\mathcal{X}|^2} \sum_{{\bf x_m}, {\bf x_n} \in \mathcal{X}}
		[\nabla_U \bar{W}_m(\theta(t))]^{\top}
		K({\bf x_m}, {\bf x_n}; \theta_0)
		[\nabla_U \bar{W}_n(\theta(t))], 
	\end{split}
\end{equation}
where 
\[
\bar{W}_n(\theta) = W({\bf x_n}, \bar{u}({\bf x_n}; \theta), \bar{u}^{\prime}({\bf x_n}; \theta))
\]
and
\[
\bar{I}_{\mathcal{X}}(\theta) = \frac{1}{|\mathcal{X}|} \sum_{{\bf x_n} \in \mathcal{X}} W({\bf x_n}, \bar{u}({\bf x_n}; \theta),  \bar{u}^{\prime}({\bf x_n}; \theta))
\]
are the linearization of the Lagrangian $W$ and the empirical loss~\eqref{eqn:ensemble-loss} at $\theta_0$, respectively, 
and 
$
K({\bf x_m}, {\bf x_n}; \theta_0)
$
is the NTK parameterized at the initial guess $\theta_0$.
It has been shown that when the minimum width of the DNN is sufficiently large, the NTK $K({\bf x}, {\bf x^{\prime}}; \theta_0)$ becomes independent of the initialization $\theta_0$~\cite{lee2019wide, arora2019exact} and we can define the asymptotic NTK (independent of the parameterization)
\begin{equation}\label{eqn:NTK-matrix}
	\bar{K}({\bf x}, {\bf x^{\prime}}) \triangleq \lim_{\text{NN width}\to \infty} \mathbb{E}_{\theta_0} \left\{K({\bf x}, {\bf x^{\prime}}; \theta_0)\right\} \in \mathbb{R}^{2 \times 2}.
\end{equation} 
Finally, we obtain the linearized loss dynamics of DRM (upon replacing $K$ 
by $\bar{K}$ and a vectorization of~\eqref{eqn:loss-trajectory-linearize})
\begin{equation}\label{eqn:loss-trajectory-NTK}
	\frac{d \bar{I}_{\mathcal{X}}(\theta(t))}{dt} 
	=
	-\frac{\eta}{|\mathcal{X}|^2}  [\nabla_{U} \bar{W}_{\mathcal{X}}(\theta(t))]^{\top} M_{\mathcal{X}} [\nabla_{U} \bar{W}_{\mathcal{X}}(\theta(t))],
\end{equation}
where the block Gram matrix
$
M_{\mathcal{X}}
$
consists of $\bar{K}({\bf x_m}, {\bf x_n})$ at its $(m, n)$-th block,
i.e., 
\begin{equation}\label{eqn:Gram-matrix}
	M_{\mathcal{X}} = \left(\bar{K}({\bf x_m}, {\bf x_n})\right)_{m,n = 1, \ldots, |\mathcal{X}|}  \in \mathbb{R}^{2|\mathcal{X}| \times 2|\mathcal{X}|},
\end{equation}
and $\bar{W}_{\mathcal{X}} = [\bar{W}_1, \ldots, \bar{W}_{|\mathcal{X}|}]^{\top} \in \mathbb{R}^{|\mathcal{X}| \times 1}$
and
$\nabla_U \bar{W}_{\mathcal{X}} = [\nabla_U \bar{W}_1, \ldots, \nabla_U \bar{W}_{|\mathcal{X}|}]^{\top} \in \mathbb{R}^{ 2|\mathcal{X}|\times 1}$.

We make two important observations from the loss dynamics~\eqref{eqn:loss-trajectory-NTK}: 1) Assuming $M_{\mathcal{X}}$ is positive definite, the convergence of the loss function $\bar{I}_{\mathcal{X}}(\theta(t))$
to a critical point is equivalent to the gradient of the Lagrangian vectors, i.e., $\nabla_{U} \bar{W}_{\mathcal{X}}(\theta(t))$, converges to zero;
2) If $\bar{I}_{\mathcal{X}}(\theta)$ is convex and bounded from below, $\theta(t)$ converges to the global minimum of $\bar{I}_{\mathcal{X}}(\theta)$.
However, the loss dynamics says nothing about the rate of convergence to a critical point. 

Therefore, we further assess the convergence rate of $\nabla_{U} \bar{W}_{\mathcal{X}}(\theta(t))$ to zero by considering its time evolution given by (derivation is postponed to~\ref{app:gradient-evolution})
\begin{equation}\label{eqn:gradient-evolution}
	\begin{split}
		\frac{d[\nabla_{U} \bar{W}_{\mathcal{X}}(\theta(t))]}{dt}
		=
		-\frac{\eta}{|\mathcal{X}|} D_{\mathcal{X}}(\theta(t)) M_{\mathcal{X}} [\nabla_{U} \bar{W}_{\mathcal{X}}(\theta(t))],
	\end{split}
\end{equation}
where the block diagonal matrix $D_{\mathcal{X}}(\theta(t))$ consists of $2\times 2$ Hessians of $\bar{W}_n \triangleq W(x_n, \bar{u}_n, \bar{u}_n^{\prime})$, i.e., 
\begin{equation*}
	D_{\mathcal{X}}(\theta(t)) = \text{diag}\left( \begin{bmatrix}
		\partial_{uu}^2 \bar{W}_n & \partial_{uu^{\prime}}^2 \bar{W}_n \\
		\partial_{u^{\prime}u}^2 \bar{W}_n & \partial_{u^{\prime}u^{\prime}}^2 \bar{W}_n
	\end{bmatrix}    \right)_{n=1, \ldots, |\mathcal{X}|} \in \mathbb{R}^{2|\mathcal{X}| \times 2|\mathcal{X}|}.
\end{equation*}
Now we are a in position to present the NTK theorem for DRM, which is a direct consequence of~\eqref{eqn:gradient-evolution}.
\begin{theorem}\label{thm:convergence}
	Suppose that 
	\begin{enumerate}
		\item the lazy training assumption (see e.g.,~\cite{chizat2019lazy}) is satisfied such that $D_{\mathcal{X}}(\theta(t)) \approx D_{\mathcal{X}} \triangleq D_{\mathcal{X}}(\theta_0)$;
		\item the Lagrangian $W$ is strictly convex in $(u, u^{\prime})$ such that the matrix $D_{\mathcal{X}}$ is positive definite;
		\item the Gram matrix $M_{\mathcal{X}}$ induced by the NTK~\eqref{eqn:NTK-matrix} is positive definite.
	\end{enumerate}
	Then, the asymptotic gradient (with respect to $u$ and $u^{\prime}$) dynamics of the Lagrangian $W$ in DRM is given by~\eqref{eqn:gradient-evolution}. 
	Moreover, we have
	\[
	[Q\nabla_{U} \bar{W}_{\mathcal{X}}(\theta(t))]^{\top} 
	=
	\text{e}^{-\eta \Lambda t /|\mathcal{X}|} [Q\nabla_{U} \bar{W}_{\mathcal{X}}(\theta_0)]^{\top},
	\]
	where we have used the spectral decomposition $D_{\mathcal{X}}M_{\mathcal{X}} = Q \Lambda Q^{\top}$ with 
	orthonormal matrix $Q = [q_1, \ldots, q_{2|\mathcal{X}|}]$ and diagonal matrix $\Lambda = \text{diag}(\lambda_1, \ldots, \lambda_{2|\mathcal{X}|})$ with $\lambda_1 \geq \lambda_2 \geq \ldots \geq \lambda_{2|\mathcal{X}|} > 0$.
\end{theorem}
A few remarks are in order.  First, the theorem suggests that the
specific convergence rate of
$\nabla_{U} \bar{W}_{\mathcal{X}}(\theta(t))$ along each direction
$q_i$ is determined by the corresponding eigenvalue $\lambda_i$. For
$\lambda_i \gg 0$, 
\[\bar{U}_{\mathcal{X}}(\theta(t)) = [\bar{U}_1(\theta(t)), \ldots, \bar{U}_{|\mathcal{X}|}(\theta(t))]^{\top} \in \mathbb{R}^{ 2|\mathcal{X}| \times 1}
\] 
with $\bar{U}_n(\theta) = [\bar{u}_n(x_n; \theta), \bar{u}_n^{\prime}(x_n; \theta)]^{\top} \in \mathbb{R}^{2\times 1}$
converges
fast along the direction $q_i$. Although for $\lambda_i \approx 0$, DNNs
have a significantly slower learning rate in the corresponding
direction $q_i$, preventing DNNs from learning the fine structure of the
solution.  
Motivated by this, we consider Fourier feature mapping to alleviate the spectrum bias issue in the next section.
Second, for a
non-convex Lagrangian $W$, the convergence of
$\nabla_{U} \bar{W}_{\mathcal{X}}(\theta(t))$ requires a more refined
analysis from variational calculus~\cite{dacorogna2007direct}, which
will be the focus of our future work. However, we empirically observed that in Section~\ref{sec:numerics} the Fourier feature mapping works equally well in the non-convex setting.
Finally, we point out that for
the type of non-convex variational problems considered in this work,
solving the corresponding Euler-Lagrange equations does not
necessarily lead to the correct minimizer and hence PINN is not
applicable.  Thus, DRM is the only option for solving variational
problem using neural networks.

\subsection{Fourier feature from the NTK perspective}
\label{subsec:FF}

\begin{figure}[!ht]
	\centering
	\includegraphics[width=1.0\textwidth]{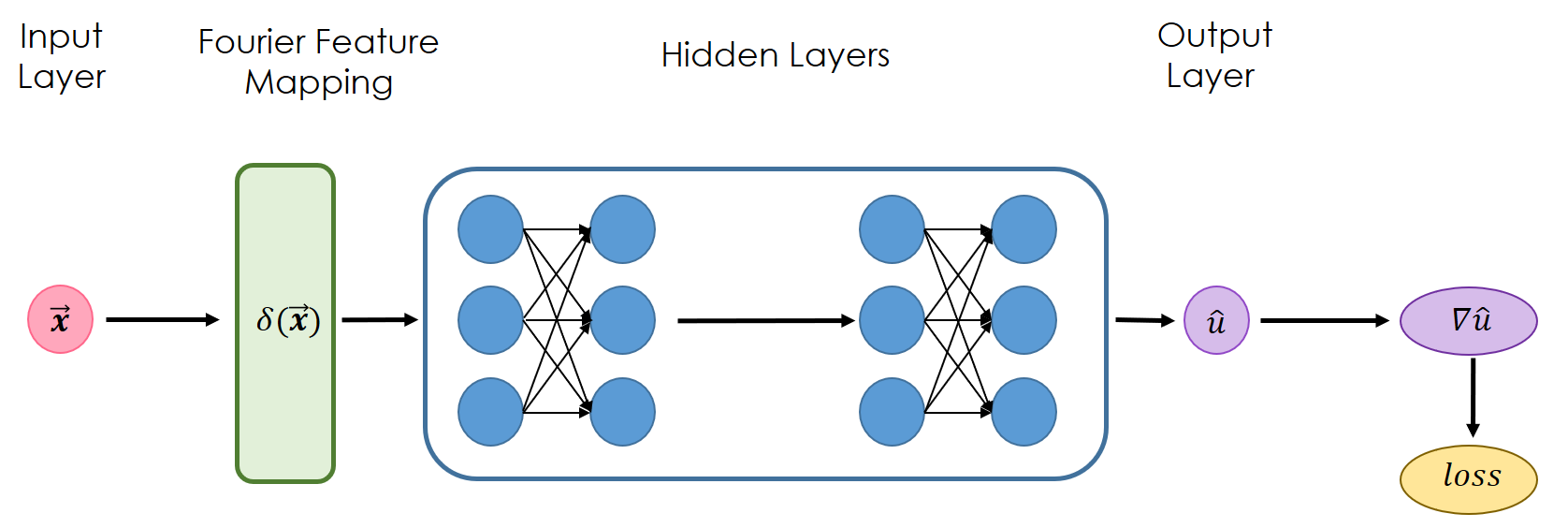}
	\caption{Structure of Neural Network by applying Fourier feature mapping to the input layer.}
	\label{fig:NN+FF_structure}
\end{figure}

To alleviate the spectral bias of DRM, we apply a Fourier feature mapping $\delta$ to the input $\bf{x}$ before it is sent to the DNN. See Figure~\ref{fig:NN+FF_structure} for the simple architecture. 
The Fourier feature mapping has been widely used in various fields in machine learning, e.g., large-scale kernel regression and deep learning~\cite{rahimi2007random, FourierFeatures2020}.
However, to the best of our knowledge, the reason why Fourier feature mapping enables DNNs to learn high frequency solutions is not well understood from a theoretical perspective.
In this section, we provide a heuristic argument from the NTK perspective to justify the application of Fourier feature mapping for DRM.
For simplicity, we consider an one dimensional problem ($d=1$) and assume that the Lagrangian $W = W(x, u)$. The Fourier feature mapping is chosen to be $\delta(x) = [\sin x, \cos x] \in \mathbb{S}^1$, where $\mathbb{S}^1$ is the unit circle in $\mathbb{R}^2$.
Viewing the pair ${\bf y} = [\sin x, \cos x] \in \mathbb{S}^1$ as the input of the DNN, the dataset $\mathcal{X}$ is mapped to $\mathcal{Y} = \delta(\mathcal{X}) \subset \mathbb{S}^1$. 
Under these assumptions, the asymptotic NTK defined in~\eqref{eqn:NTK-matrix} becomes a scalar valued positive definite kernel
\[
\bar{K}({\bf y_1}, {\bf y_2})
= \lim_{\text{NN width}\to \infty} \mathbb{E}_{\theta_0} \left\{\nabla_{\theta} \hat{u}({\bf y_1} ;\theta_0) [\nabla_{\theta}\hat{u}({\bf y_2} ;\theta_0)]^{\top}\right\}
, \qquad {\bf y_1}, {\bf y_2} \in \mathbb{S}^1
\]
and $M_{\mathcal{Y}}$ reduces to the usual Gram matrix evaluated at the input set $\mathcal{Y}$ (recall~\eqref{eqn:Gram-matrix} for definition), i.e.,
\[
M_{\mathcal{Y}} = \bar{K}(\mathcal{Y}, \mathcal{Y}).
\]
Note that the above argument can be easily generalized to the case where $W = W(x, u, u^{\prime})$ by considering a matrix valued kernel $\bar{K}$.
In~\ref{app:eigenspectrum}, we show that the $k$-th eigenvalue of $M_{\mathcal{Y}}$ is approximately proportional to the $k$-th eigenvalue of the NTK $\bar{K}$ (see~\eqref{eqn:eigen-L} for definition). Therefore, one may study the eigenvalues of $\bar{K}$ when concerned with the decay rate of the eigenvalues of $M_{\mathcal{Y}}$. 
It has been shown that (Theorem 1 in~\cite{geifman2020similarity}), when restricted to $\mathbb{S}^1$, the $k$-th eigenvalue of the NTK $\bar{K}$ scales as $\mathcal{O}(k^{-2})$, meaning that the eigenvalue of $\bar{K}$ has a quadratic decay rate. 
For multi-scale problems whose NTK spectrum exhibits multiple scales, e.g., an exponential decay rate $\mathcal{O}(\mathrm{e}^{-k})$, the Fourier feature mapping may homogenize the convergence rate along each direction $q_i$ hence
alleviating the spectral bias issue of the dynamics~\eqref{eqn:gradient-evolution}.
In Section~\ref{sec:numerics}, we empirically demonstrate the benefit of Fourier feature mapping when applied to multi-scale variational problems.


\section{Numerical Results \& Discussion}\label{sec:numerics}
We consider the following non-convex variational minimization problems: the first consists of a double well potential, $W(x) = (x^2-1)^2$ which leads to the following energy minimization problem: 
\begin{align}
	{\rm Minimize} \ I(u) = \int_0^1 (u_x^2-1)^2  \ dx  \qquad {\rm subject \ to} \qquad u(0) = u(1) = 0.
	\label{eqn:1D_Problem_1}
\end{align}

Note that the first component of the energy density is non-negative with zeros at $u_{x} = \pm 1$, which are often called zero-energy wells and correspond to the preferred phases of the problem. 
We note that for this particular problem, the minimum is attained: Carstensen showed that all Lipschitz continuous functions $u(x)$, with slope $u_{x} = \pm 1$ almost everywhere, minimize $I$ \cite{Carstensen_2005}. The energy of such function is $I = 0$. It should be emphasized that while deriving the Euler-Lagrange equation for non-convex problems like~\eqref{eqn:1D_Problem_1} is possible as shown below:
\begin{align}
	\frac{d}{dx} [u_x (u_x^2-1)] = 0 
	\label{eqn:Euler-Lagrange}
\end{align}
its solution $u(x) = 0$ does not minimize~\eqref{eqn:1D_Problem_1}. Consequently, applying the PINN algorithm to the strong form equations is not viable, as the algorithm would inevitably converge to the trivial solution.

The second benchmark problem is a variation of the double well potential, where a lower order term of the form $u^2$ is introduced, generating the following minimization problem:
\begin{align}
	{\rm Minimize} \ I(u) = \int_0^1 (u_x^2-1)^2 + u^2 \ dx \qquad {\rm subject \ to} \qquad u(0)=u(1) = 0.
	\label{eqn:1D_Problem_2}
\end{align}
We note that no minimizer exists for this problem. The infimum, although zero, cannot be attained since there is no function that satisfies $u = 0$ and $u_{x} = \pm 1$ almost everywhere. Minimizing sequences oscillate and converge weakly, but not strongly, to zero \cite{Carstensen_2005, Muller1993, Kohn&Otto}. This is the first simple  example that demonstrates how minimization can lead to fine scale oscillations or microstructure formation.

Finally, the third problem considered here is the $2D$ scalar problem for twin branching, which takes the following form:  
\begin{align}
	{\rm Minimize} \ I(u) = \int_{\Omega} u_x^2 + (u_y^2-1)^2  \ dx dy \qquad {\rm subject \ to} \qquad u=0  \ {\rm on} \ \partial \Omega,  
	\label{eqn:2D_Problem}
\end{align}
where $\Omega = [0,1]^2$. As in the previous problem, no minimizers exist since there is no function that can satisfy the integrand and boundary conditions at the same time, leading to minimizing sequences that develop rapid oscillations \cite{Muller}.

Recall that Chen \textit{et al}.\ applied DRM to non-convex energy problems in $1D$ and $2D$, similar to the ones described above. We now discuss the differences and similarities between our benchmark problems and those examined in~\cite{Chen&Rosakis2023}. Comparable to~\eqref{eqn:1D_Problem_1}, the $1D$ minimization problem in~\cite{Chen&Rosakis2023} is comprised of a double-well potential energy density subject to Dirichlet boundary conditions. Both minimization problems consist of a minimum energy ($I =0$) which can be obtained through multiple continuous functions $u(x)$, leading to loss of uniqueness. A key distinction lies in the minima locations; in~\cite{Chen&Rosakis2023}, they occur at $0$ and $1$, while in ~\eqref{eqn:1D_Problem_1}, they occur at $-1$ and $1$. Our Dirichlet boundary conditions are fixed at $0$, contrasting with~\cite{Chen&Rosakis2023} where the left boundary is fixed at $0$ and the right boundary is fixed at $\gamma$ where $\gamma \in \mathbb{R}$. This leads to solutions with slopes $u_x = 0$ and $1$ in~\cite{Chen&Rosakis2023}, whereas the solutions to~\eqref{eqn:1D_Problem_1} have slopes $u_x = \pm 1$.

Similarly, the $2D$ minimization problem in~\cite{Chen&Rosakis2023} mirrors features found in~\eqref{eqn:2D_Problem}. Both problems consist of a double well energy potential and are subject to Dirichlet boundary conditions, which yield to minimizing sequences with rapid oscillations but no actual minimizers. The main differences between~\eqref{eqn:2D_Problem} and the 2D problem in~\cite{Chen&Rosakis2023} lie in the minima locations of the energy well potential ($(\pm 1,0)$ in~\eqref{eqn:2D_Problem} vs. $(0,0)$ and $(1,0)$ in~\cite{Chen&Rosakis2023}). The Dirichlet boundary conditions are set to $0$ across the boundary in~\eqref{eqn:2D_Problem}, while Chen \textit{et al}.\ set $u(x,y) = \gamma x$ with $\gamma \in \mathbb{R}$ in~\cite{Chen&Rosakis2023}.

Given that the distinctions mentioned above are cosmetic and do not alter the fundamental structure of the minimization problems, we anticipate the hypothesis and conclusions articulated in \cite{Chen&Rosakis2023}, particularly the hypothesis that increasing the DNN increases the number of twin bands for the $2D$ problem, remain true for~\eqref{eqn:2D_Problem}. We test this hypothesis numerically in the sections below.

\label{sec:Numerics}
\subsection{$1D$ Benchmark Problem \# 1}\label{sec:1D_results_1}
We start our discussion by approximating the solution to~\eqref{eqn:1D_Problem_1} using DRM without Fourier mapping (as described in~\cite{Chen&Rosakis2023}) and compare the results with the new algorithm: DRM with Fourier mapping (DRM\&FM). In both cases, a fully connected feed-forward neural network with an input layer, multiple hidden layers and an output layer is constructed. The input layer consists of one node (for the $x$ coordinate of our problem),  each hidden layer consists of $128$ nodes, and the output layer consists of one node  (used to output the approximated solution $\hat{u}$). Consistent with~\cite{Chen&Rosakis2023}, we apply the ReLU activation function in each layer.  To accelerate training, we use Adams Optimizer on a mini-batch size of $128$ collocation points sampled uniformly, with an initial learning rate $\eta = 10^{-4}$. We implement a cosine annealing schedule that decreases the learning rate to zero over the course of the simulation. The boundary conditions are enforced using the penalty approach with a penalty parameter set to $\lambda = 500$.

\begin{figure}[!ht]
	\centering
	\subfigure[~]{\includegraphics[width=0.31\textwidth]{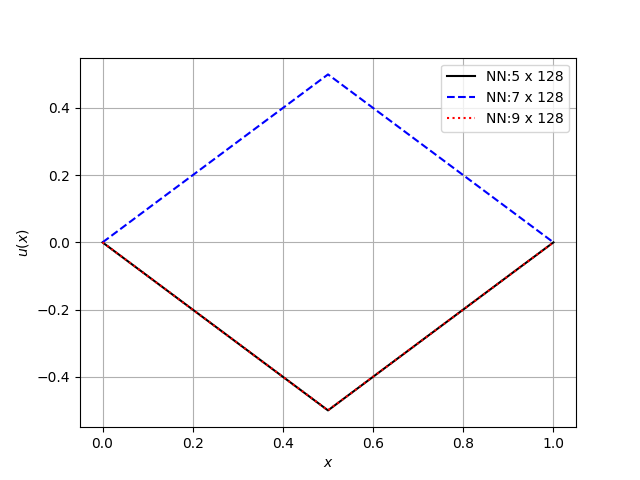}\label{subfig:u_1D_no_FF}}
	\subfigure[~]{\includegraphics[width=0.31\textwidth]{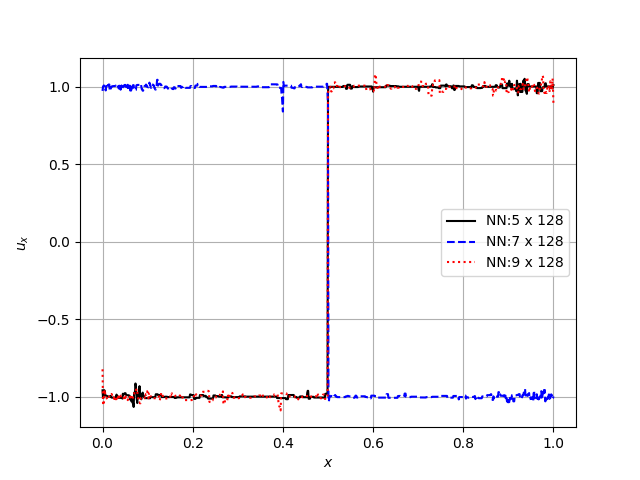}\label{subfig:ux_1D_no_FF}}
	\caption{(a)~DRM approximation to~\eqref{eqn:1D_Problem_1} with ReLU activation function, $\eta = 1.0\times 10^{-4}$ after $100000$ epochs with DNN structure of $5$, $7$ and $9$ hidden layers. (b)~ The derivative $u_x$ of the DRM approximation to~\eqref{eqn:1D_Problem_1}.}
	\label{fig:1D_problem_noFF}
\end{figure}

Recall that there exist multiple solutions that minimize~\eqref{eqn:1D_Problem_1}: namely, any function $u(x)$ with slope $u_x = \pm 1$ almost everywhere minimizes the functional $I(u)$. In Figure~\ref{fig:1D_problem_noFF} we present the minimizing solutions generated by DRM with no Fourier mapping as we vary the depth of the network while setting the learning rate initially to $\eta = 10^{-4}$.  We see that for this particular benchmark problem, increasing the depth of the DNN does not generate high-frequency solutions, analogous to the increased number of twin bands of the 2D problem discussed in~\cite{Chen&Rosakis2023}. Solutions with one transition between the two preferred interfaces ($u_x = \pm 1$) are generated for a DNN with $5$, $7$ and $9$ hidden layers (see Fig~\ref{subfig:ux_1D_no_FF}). 

\begin{figure}[ht!]
	\centering
	\subfigure[$i = 2$]{\includegraphics[width=0.31\textwidth]{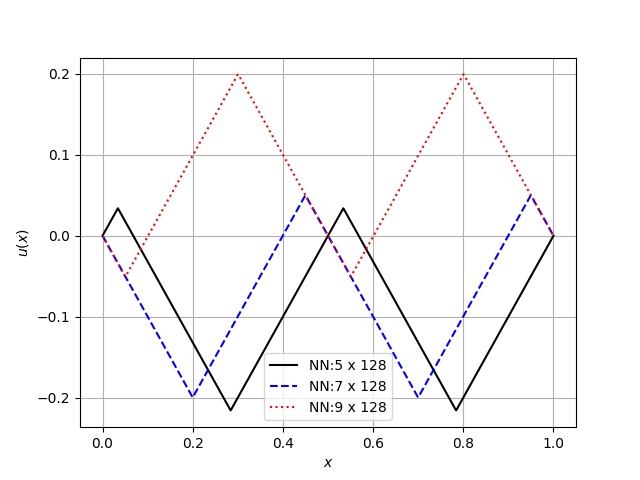}\label{fig:u_1D_FF=2}}
	\subfigure[$i = 3$]{\includegraphics[width=0.31\textwidth]{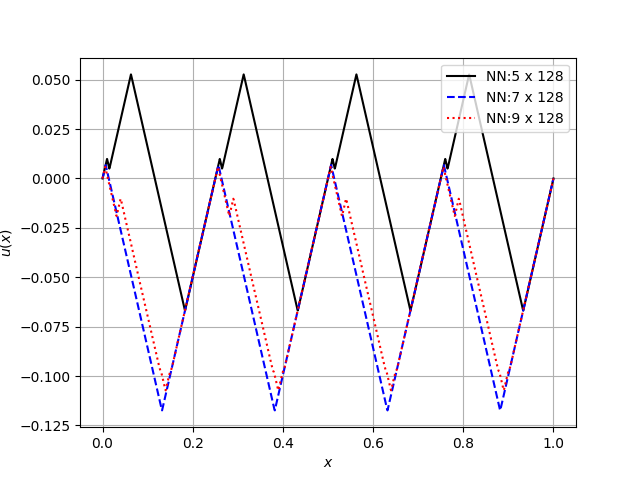}\label{fig:u_1D_FF=3}}
	\subfigure[$i = 4$]{\includegraphics[width=0.31\textwidth]{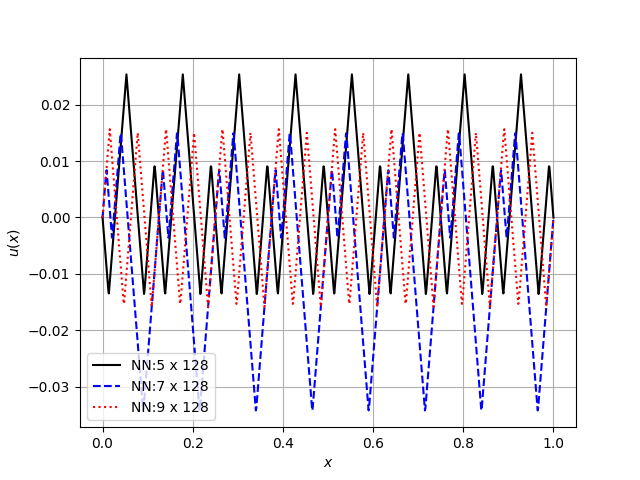}\label{fig:u_1D_FF=4}}
	\caption{DRM approximation to~\eqref{eqn:1D_Problem_1} where a NN with $5$,$7$ and $9$-hidden layers, ReLU activation function, $\eta = 1.0\times 10^{-4}$ and Fourier mapping of frequency $\delta({\bf x}) = \left[\sin(2^i\pi {\bf x}),\cos(2^i\pi {\bf x})\right] $ after $100000$ epochs. }
	\label{fig:1D_problem_FF}
\end{figure}

Figure~\ref{fig:1D_problem_FF} displays the solution generated by DRM with Fourier feature mapping under the same conditions. Recall that the information passes from the input layer, through a Fourier mapping of the form $\delta({\bf x}) = \left[\sin(2^i\pi {\bf x}), \cos(2^i \pi {\bf x})\right]$ with $i = 2, 3, 4$ and $\bf{x} \in \mathbb{R}$, to the hidden and output layers. 
We observe that the frequency of the mapping can be leveraged to generate minimizing solutions of high frequency, independently of the depth of the DNN.  
When passing a Fourier mapping of frequency $4\pi$ as shown in Fig.~\ref{fig:u_1D_FF=2} ($8\pi$ as shown in Fig.~\ref{fig:u_1D_FF=3}), we generate a solution with $4$ ($8$) transitions between preferred states, independently of the depth of the DNN. When applying a Fourier mapping of frequency $16\pi$ however, we get mixed results: implementing a DNN with $5$ and $7$ hidden layers leads to a solution with $32$ transitions between states (as shown by the black and red dotted lines in Fig.~\ref{fig:u_1D_FF=4}), while a $9$ layer DNN leads to a solution with $16$ transitions between preferred states (as shown by blue dashed lines). Figure~\ref{fig:1D_problem_FF} shows that increasing the frequency of the Fourier mapping increases the number of transitions between the preferred states but one cannot quantify the relationship between mapping frequency and number of transitions within the domain.

\subsection{$1D$ Benchmark Problem \# 2}\label{sec:1D_results_2}
\begin{figure}[ht!]
	\centering
	\subfigure[NN: $3\times 128$]{\includegraphics[width=0.31\textwidth]{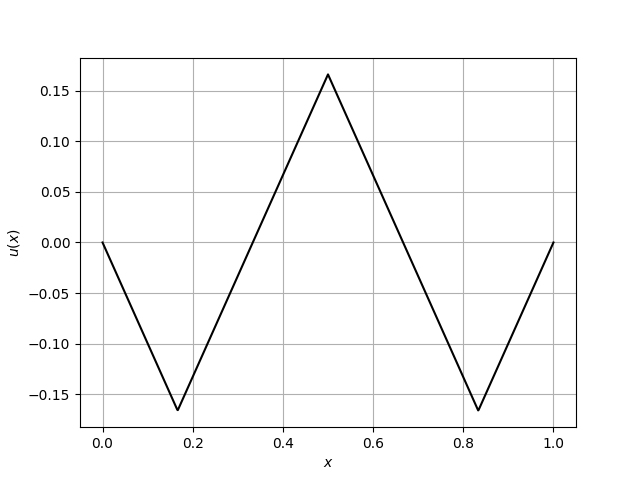}\label{subfig:u2_1D_noFF_d=3_200K}}
	\subfigure[NN: $5\times 128$]{\includegraphics[width=0.31\textwidth]{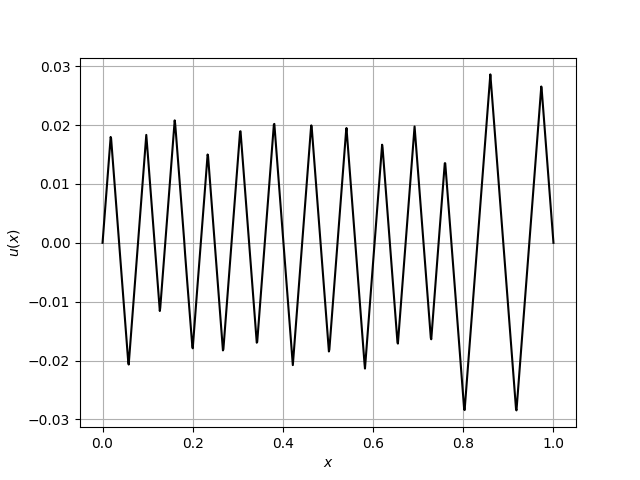}\label{subfig:u2_1D_noFF_d=5_200K}}
	\subfigure[NN: $7\times 128$]{\includegraphics[width=0.31\textwidth]{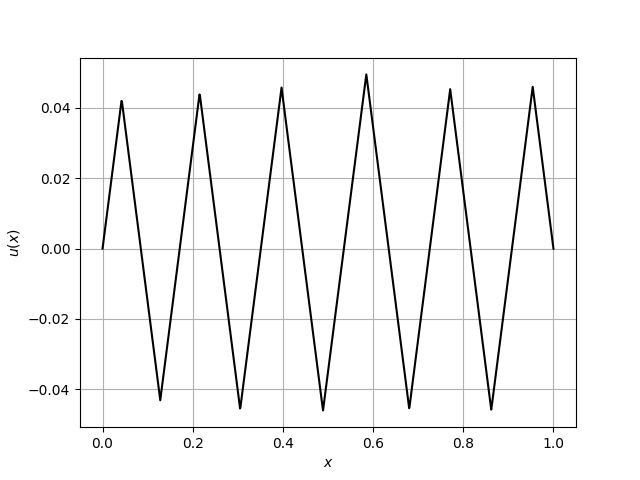}\label{subfig:u2_1D_noFF_d=7_200K}}\\
	\subfigure[NN: $3\times 128$]{\includegraphics[width=0.31\textwidth]{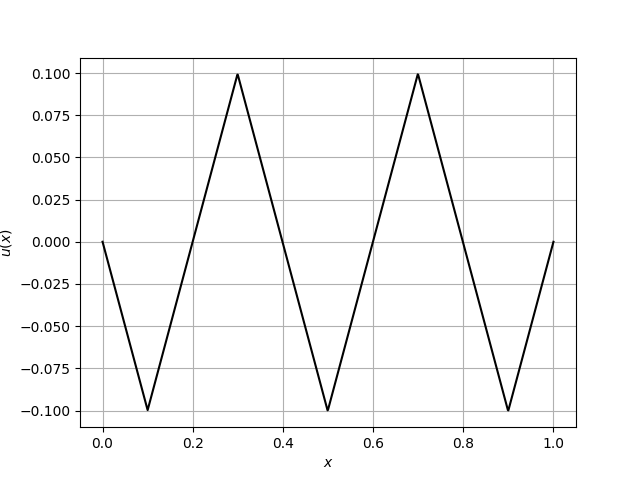}\label{subfig:u2_1D_noFF_d=3_500K}}
	\subfigure[NN: $5\times 128$]{\includegraphics[width=0.31\textwidth]{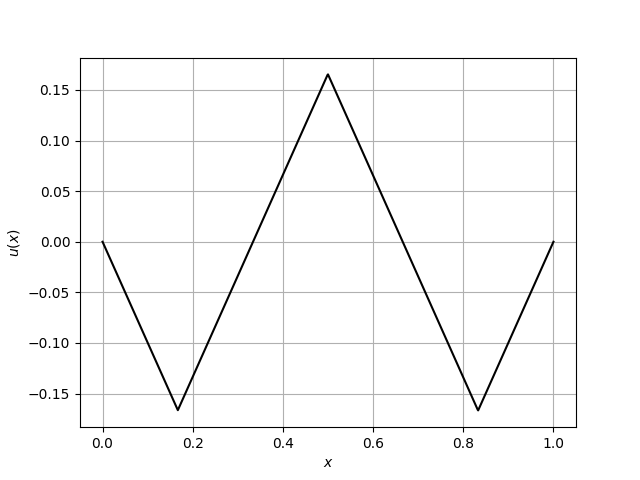}\label{subfig:u2_1D_noFF_d=5_500K}}
	\subfigure[NN: $7\times 128$]{\includegraphics[width=0.31\textwidth]{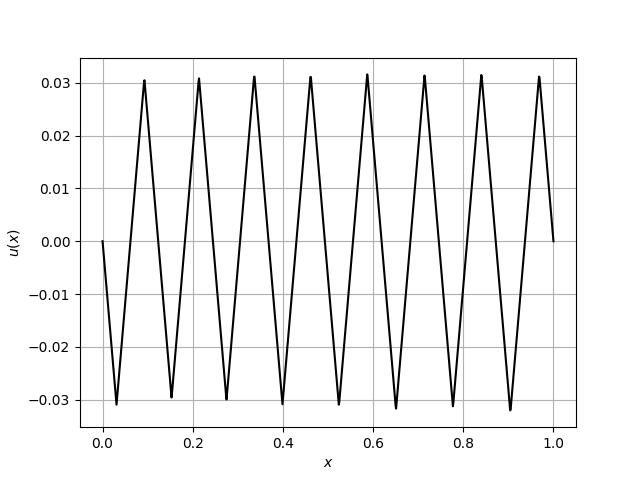}\label{subfig:u2_1D_noFF_d=7_500K}}
	\caption{{\bf First Row (a)-(c)}: DRM approximation to~\eqref{eqn:1D_Problem_2} with ReLU activation function, $\varepsilon = 0$, $\eta = 1.0\times 10^{-4}$ and cosine annealing after $200000$ epochs.
		{\bf Second Row (d)-(f):} Row: DRM approximation to~\eqref{eqn:1D_Problem_2} with ReLU activation function, $\varepsilon = 0$, $\eta = 1.0\times 10^{-4}$ and cosine annealing after $500000$ epochs. }
	\label{fig:1D_problem_2_noFF}
\end{figure}

We now discuss how DRM alone and DRM with Fourier mapping (DRM\&FM) approximate the solution sequences to the second benchmark problem given by~\eqref{eqn:1D_Problem_2}. Recall that no minimizer exists for this problem since there is no function that satisfies the conditions $u = 0$ and $u_x = \pm 1$ everywhere. Figure~\ref{fig:1D_problem_2_noFF} shows the DNN approximation of the minimizing solution to~\eqref{eqn:1D_Problem_2} as the depth of the DNN increases with no Fourier mapping after $200,000$ epochs (First Row) and $500,000$ epochs (Second Row). We observe that increasing the depth of DNN does not consistently increase the number of transitions between the preferred states. Increasing the depth of the DNN from $3$ to $5$ hidden layers increases the number of transitions for $200,000$ epochs. However, the number of transitions decreases as the depth of the DNN is increased from $5$ to $7$ hidden layers. A similar occurrence can be observed in the second row of Fig.~\ref{fig:1D_problem_2_noFF} where our simulations are run for $500,000$ epochs. In this case, increasing the depth of the DNN from $3$ to $5$ hidden layers decreased the number of transitions while increasing the depth from $5$ to $7$ hidden layers increased the number of transitions between preferred states. Based on our simulations, we can say that increasing the depth of the DNN does not consistently generate high-frequency solutions for the $1D$ benchmark problem given by~\eqref{eqn:1D_Problem_2}. We also note that a DNN with 7 hidden layers run for $500,000$ epochs was able to generate a minimizing sequence with $16$ transitions between preferred states. 

\begin{figure}[ht!]
	\centering
	\subfigure[$i = 1$]{\includegraphics[width=0.31\textwidth]{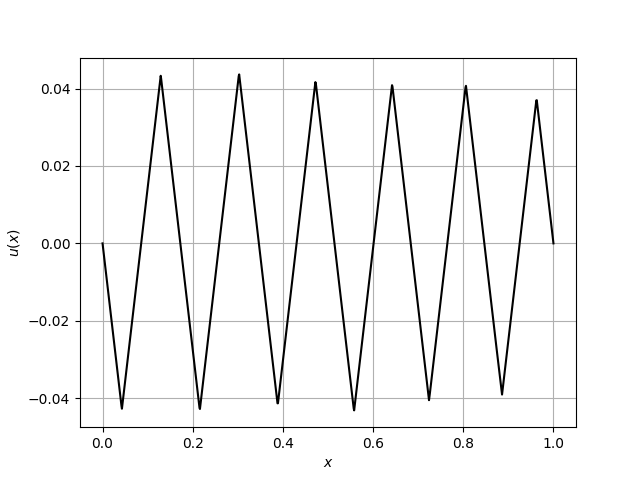}\label{subfig:u2_1D_FF=1_200K}}
	\subfigure[$i = 2$]{\includegraphics[width=0.31\textwidth]{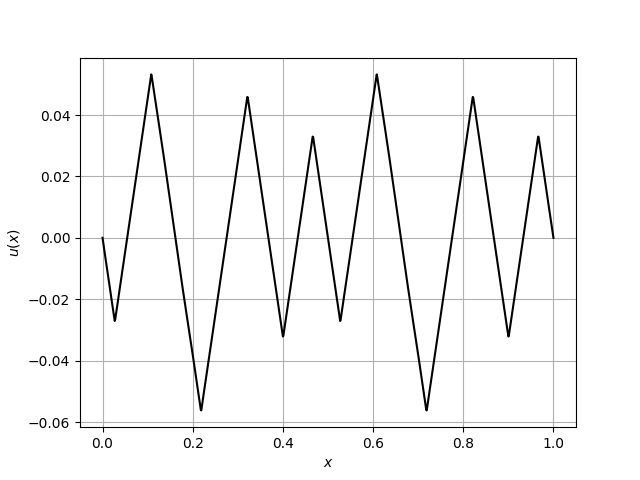}\label{subfig:u2_1D_FF=2_200K}}
	\subfigure[$i = 3$]{\includegraphics[width=0.31\textwidth]{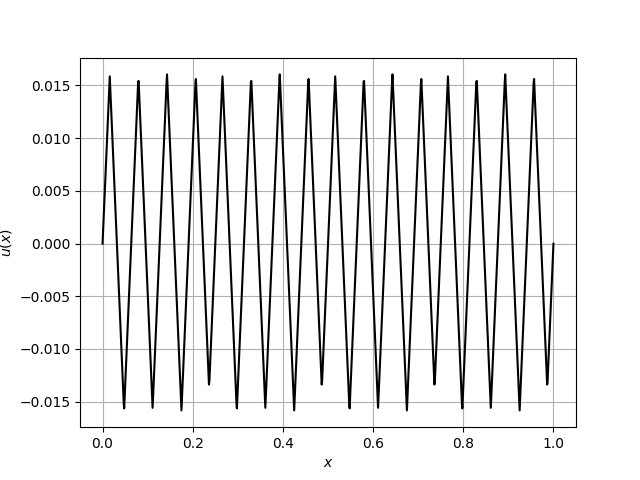}\label{subfig:u2_1D_FF=3_200K}}\\
	\subfigure[$i = 1$]{\includegraphics[width=0.31\textwidth]{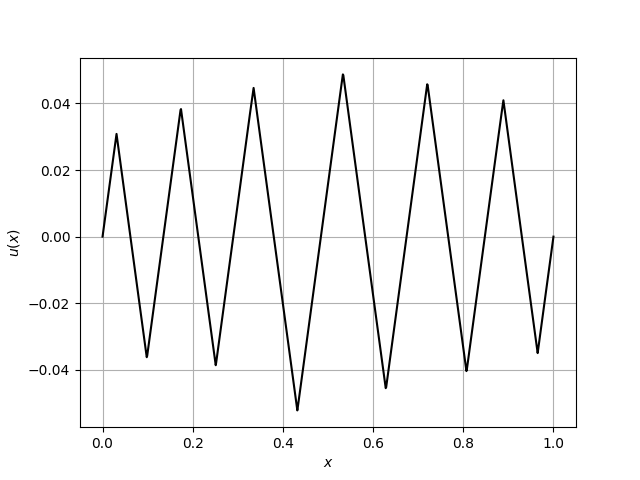}\label{subfig:u2_1D_FF=1_500K}}
	\subfigure[$i = 2$]{\includegraphics[width=0.31\textwidth]{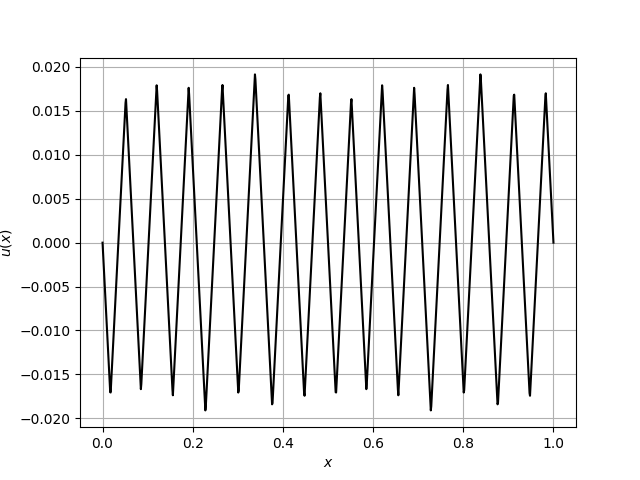}\label{subfig:u2_1D_FF=2_500K}}
	\subfigure[$i = 3$]{\includegraphics[width=0.31\textwidth]{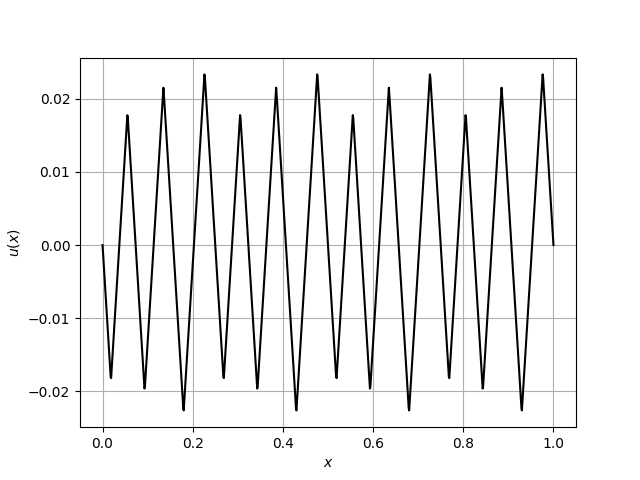}\label{subfig:u2_1D_FF=3_500K}}    
	\caption{{\bf First Row (a)-(c):} DRM\&FM approximation to~\eqref{eqn:1D_Problem_2} with $3 \times 128$ NN (3 hidden layers), ReLU activation function, $\varepsilon = 0$, $\eta = 1.0\times 10^{-4}$ and Fourier feature of frequency $\delta({\bf x}) = \left[\sin(2^i\pi {\bf x}),\cos(2^i\pi {\bf x})\right] $ after $200000$ epochs. {\bf Second Row (d)-(f):} DRM\&FM approximation under the same conditions after $500,000$ epochs. }
	\label{fig:1D_problem_2_FF}
\end{figure}

Figure~\ref{fig:1D_problem_2_FF} shows the minimizing solutions that are obtained by the DRM with a DNN structure of $3$ hidden layers and Fourier mapping of frequency $2\pi$, $4\pi$ and $8\pi$  after $200,000$ and $500,000$ epochs. We see here that the Fourier mapping with frequency $2\pi$ as shown in Figs~\ref{subfig:u2_1D_FF=1_200K} and~\ref{subfig:u2_1D_FF=1_500K} enables us to generate a solution of $12$ transitions between the two preferred states, a result that is comparable with the DRM approximation solution of a DNN of $7$ hidden layers as shown in Figs.~\ref{subfig:u2_1D_noFF_d=7_200K} \& \ref{subfig:u2_1D_noFF_d=7_500K}. Note the solution approximation consists of $11$ transitions for $200,000$ epochs and $16$ transitions for $500,000$ epochs. We note that DRM\&FM enables us to keep the number of hidden layers in the DNN fixed and generate minimizing solutions with more transitions, such as the ones shown in Fig.~\ref{fig:1D_problem_2_FF}. While it seems that the number of transitions between preferred states increases with the frequency of the Fourier mapping, the authors did not investigate the relationship between the frequency of the Fourier mapping and the number of transitions within the solution for this $1D$ problem. 

\subsection{$2D$ Benchmark Problem}\label{sec:2D_results}
We now turn to the $2D$ twin branching problem given by~\eqref{eqn:2D_Problem} and investigate whether the DRM\&FM method can be extended to generate solutions to $2D$ microstructure problems. Recall that, similar to~\eqref{eqn:1D_Problem_2}, this problem does not have a minimizer since there are no functions that can minimize the integrand and satisfy the Dirichlet boundary conditions at the same time, leading to microstructure behavior. The ideal minimizer would be a function  $u(x,y)$ such that $u_y = \pm 1$, $u_x = 0$ in $\Omega$ and $u = 0$ on $\partial \Omega$. Such function does not exist, leading to minimizing sequences with fine scale oscillations instead. 
We attempt to capture these minimizing sequences using a DNN similar in structure to the ones implemented in Secs.~\ref{sec:1D_results_1} \& \ref{sec:1D_results_2}. We adapt the DNN to minimize the $2D$ problem in~\eqref{eqn:2D_Problem} through the following changes: the input layer consists of two nodes, one for each coordinate $x$ and $y$ of our $2D$ domain,  the activation function used is of the form 
$\sigma(x) = \sqrt{x^2 + \rho^2}$, where $\rho = 0.1.$ This activation function is a variation of the SmReLU activation function used in~\cite{Chen&Rosakis2023} to better suit the problem considered here.
The DRM is run with Adams Optimizer for $300,000$ epochs with a total number of $N = 1000$ collocation points sampled uniformly across the domain ($N_{int} = 600$ in the interior  and N$_b = 400$: $100$ uniformly sampled points across each boundary). Note that we set the initial learning rate to $\eta = 10^{-4}$ and apply cosine annealing as in the 1D case. 

\begin{figure}[ht!]
	\centering
	\subfigure[NN: $3 \times 128$]{\includegraphics[width=0.31\textwidth]{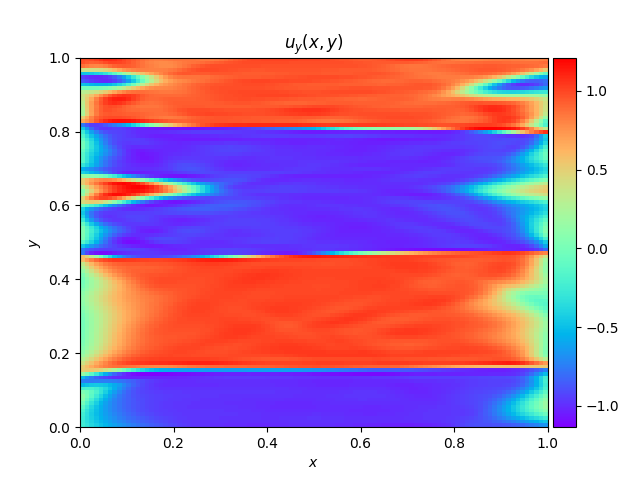}\label{fig:uy_2D_no_FF_5_eps=0}\label{subfig:uy_no_FF_depth=3}}
	\subfigure[NN: $5 \times 128$]{\includegraphics[width=0.31\textwidth]{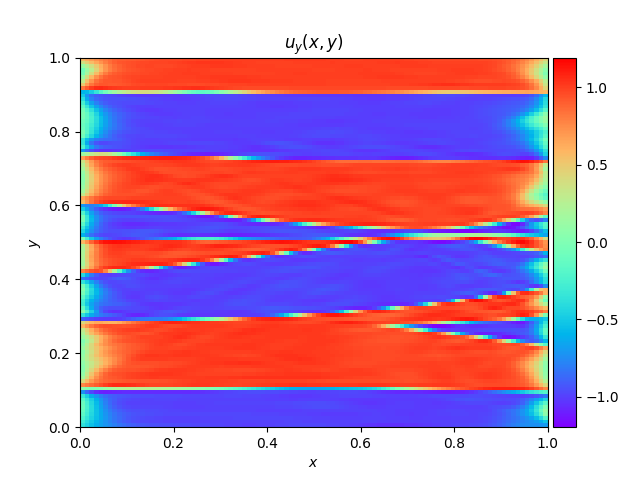}\label{fig:uy_2D_no_FF_7_eps=0}\label{subfig:uy_no_FF_depth=5}}
	\subfigure[NN: $7 \times 128$]{\includegraphics[width=0.31\textwidth]{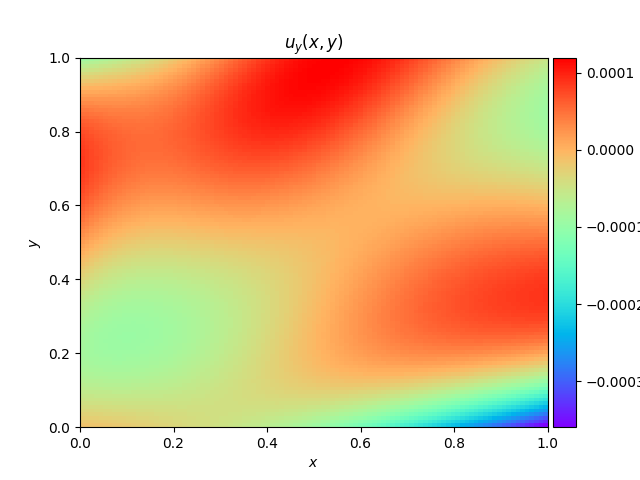}\label{fig:uy_2D_no_FF_9_eps=0}\label{subfig:uy_no_FF_depth=7}}
	\caption{DRM approximation to~\eqref{eqn:2D_Problem} with activation function $\sigma(x) = \sqrt{x^2+\rho^2}, \rho = 0.1$, $\eta = 1.0\times 10^{-4}$ and no Fourier Feature after $300000$ epochs.}
	\label{fig:2D_problem_no_FF_eps=0.0}
\end{figure}

Figure~\ref{fig:2D_problem_no_FF_eps=0.0} displays the minimizing sequences to~\eqref{eqn:2D_Problem} (we plot $u_y$ instead of $u$ to show the transition between the two preferred states $u_y = \pm 1$) as we increase the number of hidden layers in the DNN. Here, as in Sec.~\ref{sec:1D_results_1}, we consider a DNN with $3$, $5$ and $7$ hidden layers respectively and no Fourier mapping. We see that as the depth of the DNN increases, the number of bands stays the same. In fact, for a network with $7$ hidden layers, the solution is stuck to an unstable state ($u = 0$). We note that for this particular problem, increasing the depth of the DNN does not generate minimizing sequences with a large number of twin bands (high frequency). It seems like the depth of the NN is hindering the DNN from converging to a minimum: instead, it is stuck at a saddle point in the energy density functional of~\eqref{eqn:2D_Problem}.

\begin{figure}[ht!]
	\centering
	\subfigure[$i = 1$]{\includegraphics[width=0.31\textwidth]{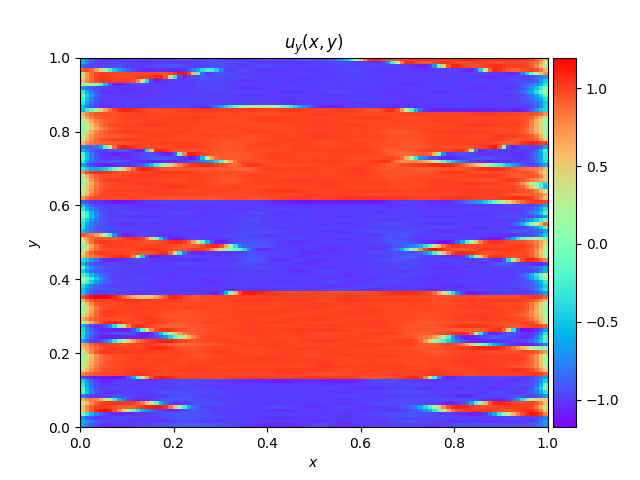}\label{subfig:uy_FF=1_epsbar=0.0}}
	\subfigure[$i = 2$]{\includegraphics[width=0.31\textwidth]{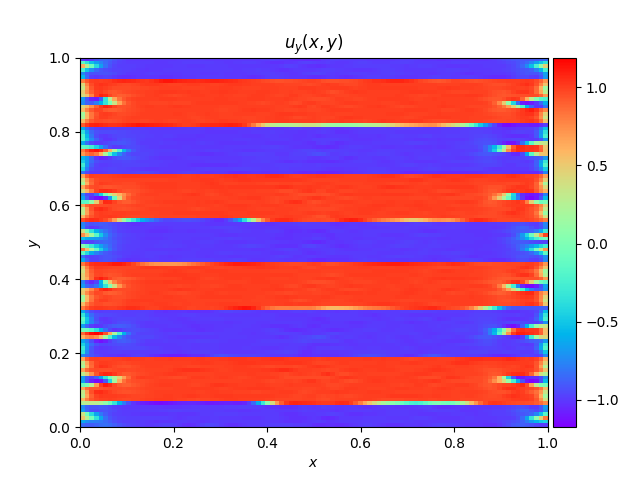}\label{subfig:uy_FF=2_epsbar=0.0}}
	\subfigure[$i = 3$]{\includegraphics[width=0.31\textwidth]{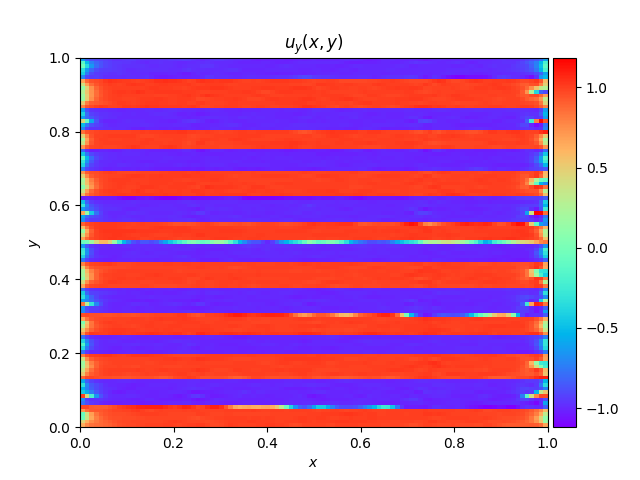}\label{subfig:uy_FF=3_epsbar=0.0}}
	\subfigure[$i = 4$]{\includegraphics[width=0.31\textwidth]{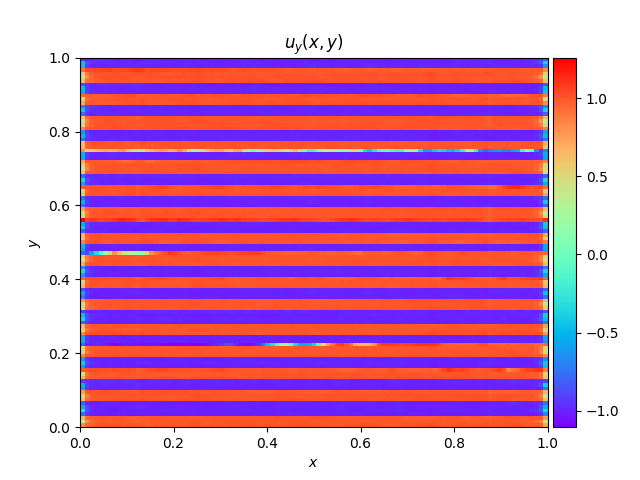}\label{subfig:uy_FF=4_epsbar=0.0}}
	\caption{DRM approximation to~\eqref{eqn:2D_Problem} with $3 \times 128$ NN and Fourier feature of frequency $\delta({\bf x}) = \left[{\bf x},\sin(2^i\pi {\bf x}),\cos(2^i\pi {\bf x})\right]$ with $\eta = 1.0\times 10^{-4}$ after $300000$ epochs.}
	\label{fig:2D_problem_FF_eps=0.0}
\end{figure}

In contrast, when a Fourier mapping of the form $\delta({\bf x}) = \left[{\bf x}, \sin(2^i\pi {\bf x}),\cos(2^i\pi {\bf x})\right]$ where $i = 1-4$  and ${\bf x} \in \mathbb{R}^2$ is applied, the number of transitions between preferred states in $u_y$ (or number of twin bands as described in~\cite{Chen&Rosakis2023}) increase (see Figure~\ref{fig:2D_problem_FF_eps=0.0}). Note that we modify the Fourier mapping by including ${\bf x}$ because a periodic solution is no longer a minimizer of the problem and we no longer expect a periodic solution in the domain. We hypothesize that applying a Fourier mapping of any frequency allows the DRM to converge to a minimizing sequence quicker than if no Fourier mapping was applied (compare Figs.~\ref{subfig:uy_FF=1_epsbar=0.0}-\ref{subfig:uy_FF=4_epsbar=0.0} with Fig.~\ref{subfig:uy_no_FF_depth=3}). We observe needle like structures forming around $x = 0$ and $x = 1$ when a Fourier mapping of low frequency is applied (see Fig.~\ref{subfig:uy_FF=1_epsbar=0.0}) but these needles do not fully grow to form additional bands in the course of our simulation. A similar behavior can be observed in Figs.~\ref{subfig:uy_FF=2_epsbar=0.0}-\ref{subfig:uy_FF=4_epsbar=0.0}: needle like structures are formed around $x = 0,1$ but these structures get smaller as the frequency of the Fourier mapping increases. Additionally, we observe that the number of twin bands increases as the frequency of the Fourier mapping increases: there are $4$ transitions between states when the frequency is set to $2 \pi$, $8$ transitions when the frequency is $4\pi$, $15$ transitions when the frequency is $8\pi$ and $32$ transitions when the frequency is $16\pi$ (See Figs.~\ref{subfig:uy_FF=2_epsbar=0.0}-\ref{subfig:uy_FF=4_epsbar=0.0}).
We observe that the minimizing solutions are noisy as the Fourier frequency increases and we attribute this noise to the fact that ~\eqref{eqn:2D_Problem} has no minimum. We emphasize that incorporating Fourier feature mapping into the DRM does not alter the number of collocation points used in the simulations ($N=1000$ in 2D case and $N = 128$ in 1D). This approach stands in sharp contrast to traditional methods like FEM, which depend heavily on mesh-size refinement to resolve the microstructure. 

\subsection{Regularized 2D Problem \& Fourier Mapping}\label{sec:2D_problem_epsilon}
Regularization is frequently used to ensure the existence of solutions to nonconvex minimization problems while also determining the length scale and fine geometry of the resulting microstructures~\cite{Muller1993, Kohn&Otto, Muller, Kohn1992}. This is achieved by adding a high-gradient term to the energy density  $W$ in~\eqref{eqn:variationalproblem}. Traditional numerical methods leverage this approach to identify the microstructure's length scale~\cite{Muller1993} and predict specific microstructure dynamics~\cite{Dondl2016}.
In this context, we consider the regularized 2D problem:
\begin{align}
	{\rm Minimize} \ I(u) = \int_{\Omega} u_x^2 + (u_y^2-1)^2 + \varepsilon^2 u_{yy}^2 \ dx dy \qquad {\rm subject \ to} \qquad u=0  \ {\rm on} \ \partial \Omega,  
	\label{eqn:2D_Problem_reg}
\end{align}
and investigate how Fourier mapping and the regularization term interact in generating high-frequency solutions to the regularized minimization problem in 2D. Recall that  $u_x$ prefers to be $0$ while $u_y$ jumps between $\pm 1$. The additional term $\varepsilon^2 u_{yy}^2$ in~\eqref{eqn:2D_Problem_reg} penalizes these transitions, facilitating the formation of fine structures by reducing the surface energy associated with the high-gradient contributions~\cite{Kohn1992}.

\begin{figure}[ht!]
	\centering
	\subfigure[no FF]{\includegraphics[width=0.31\textwidth]{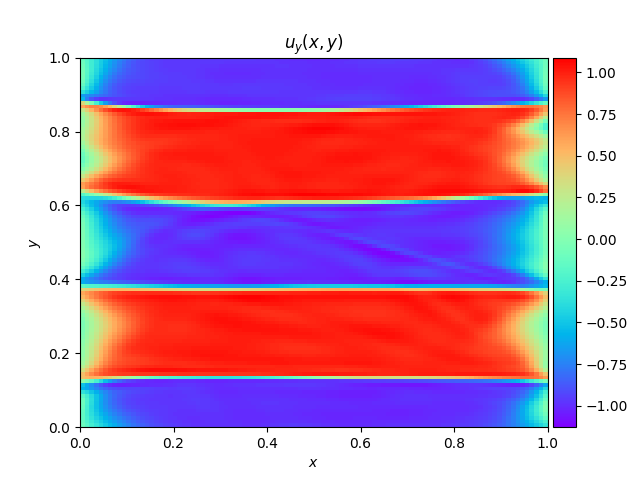}\label{subfig:uy_no_FF_eps=0.1by16}}
	\subfigure[$i = 1$]{\includegraphics[width=0.31\textwidth]{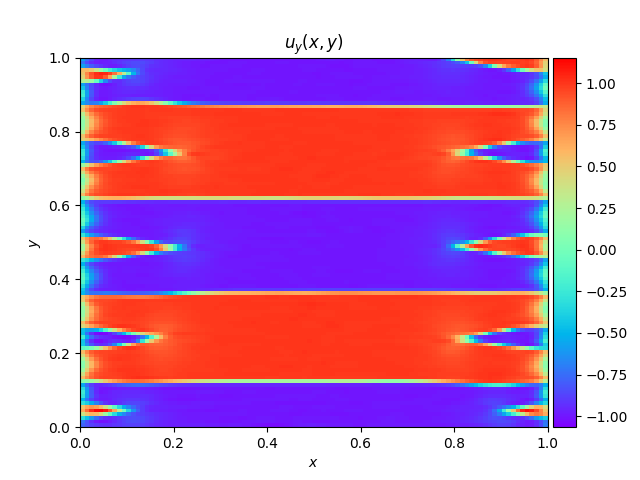}\label{subfig:uy_FF=1_eps=0.1by16}}
	\subfigure[$i = 2$]{\includegraphics[width=0.31\textwidth]{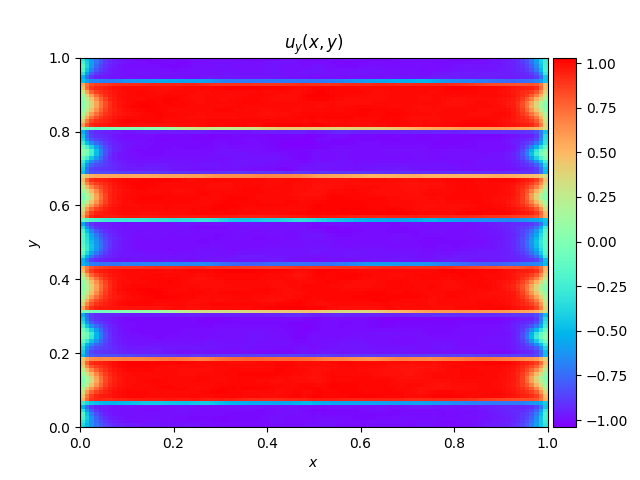}\label{subfig:uy_FF=2_eps=0.1by16}}\\
	\subfigure[$i = 3$]{\includegraphics[width=0.31\textwidth]{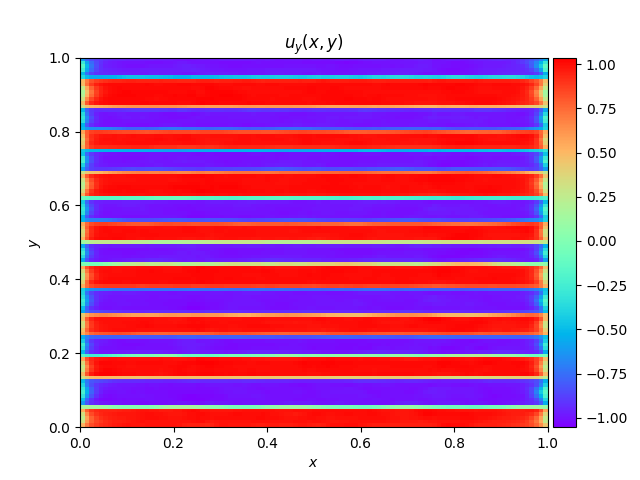}\label{subfig:uy_FF=3_eps=0.1by16}}
	\subfigure[$i = 4$]{\includegraphics[width=0.31\textwidth]{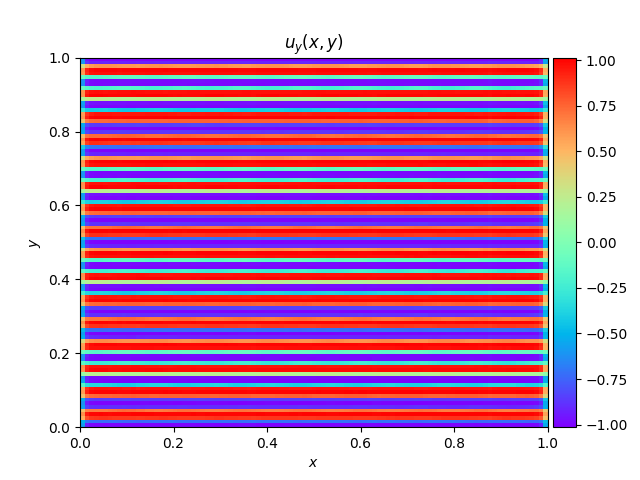}\label{subfig:uy_FF=4_eps=0.1by16}}
	\caption{DRM approximation to~\eqref{eqn:2D_Problem} with $3 \times 128$ NN and Fourier feature of frequency $\delta({\bf x}) = \left[{\bf x}, \sin(2^i\pi {\bf x}),\cos(2^i\pi {\bf x})\right]$. The activation function used is $\sigma(x) = \sqrt{x^2+\rho^2}$ with $\rho = 0.1$, $\varepsilon = 0.1/16$, $\eta = 1.0\times 10^{-4}$ after $300000$ epochs.}
	\label{fig:2D_problem_FF_eps=0.1by16}
\end{figure}

Figure~\ref{fig:2D_problem_FF_eps=0.1by16} shows the graph of the DRM generated solutions ($u_y$ instead of $u$) when $\varepsilon = 0.1/16$. We see that introducing a regularization term generates smooth minimizing sequences throughout the domain independently of whether a Fourier mapping is applied, though the Fourier mapping enables the method to generate solutions with more twin bands for large frequencies.  Comparing Figs.~\ref{subfig:uy_no_FF_depth=3} and ~\ref{fig:2D_problem_FF_eps=0.0} with Fig.~\ref{fig:2D_problem_FF_eps=0.1by16}, we observe that the regularization term helps the DRM generate smooth and symmetric solutions with smoother interfacial transitions and uniform microstructure length scales.  

\begin{figure}[ht!]
	\centering
	\subfigure[no FF]{\includegraphics[width=0.31\textwidth]{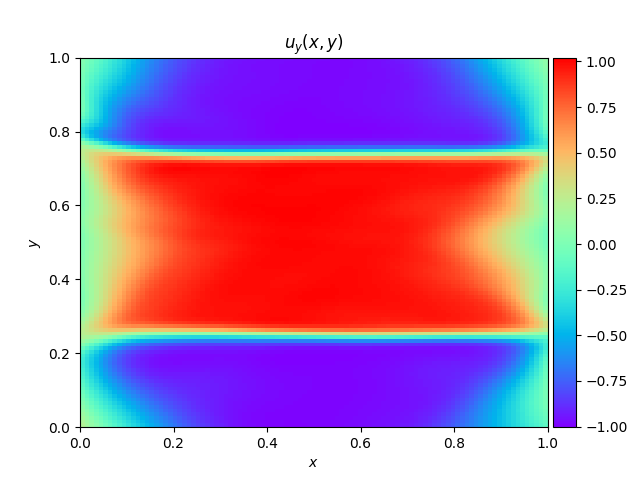}\label{subfig:uy_no_FF_eps=0.1by4}}
	\subfigure[$i = 1$]{\includegraphics[width=0.31\textwidth]{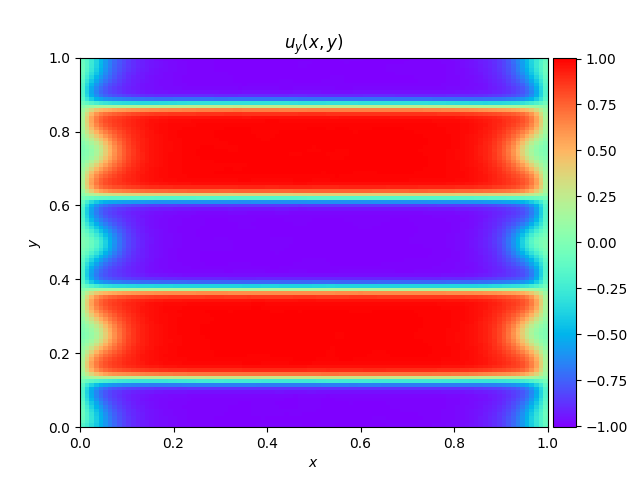}\label{subfig:uy_FF=1_eps=0.1by4}}
	\subfigure[$i = 2$]{\includegraphics[width=0.31\textwidth]{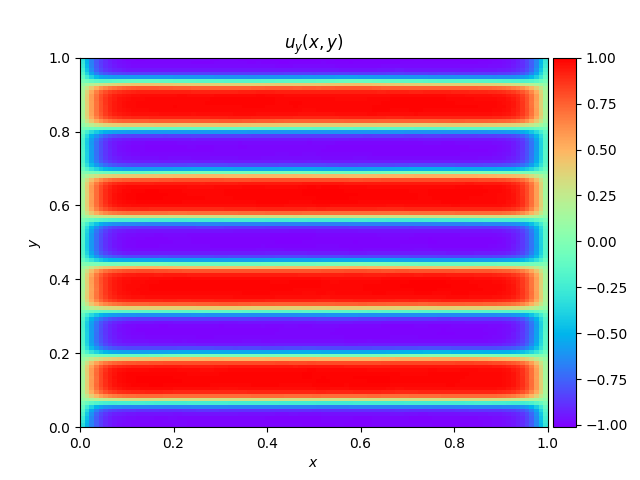}\label{subfig:uy_FF=2_eps=0.1by4}}
	\subfigure[$i = 3$]{\includegraphics[width=0.31\textwidth]{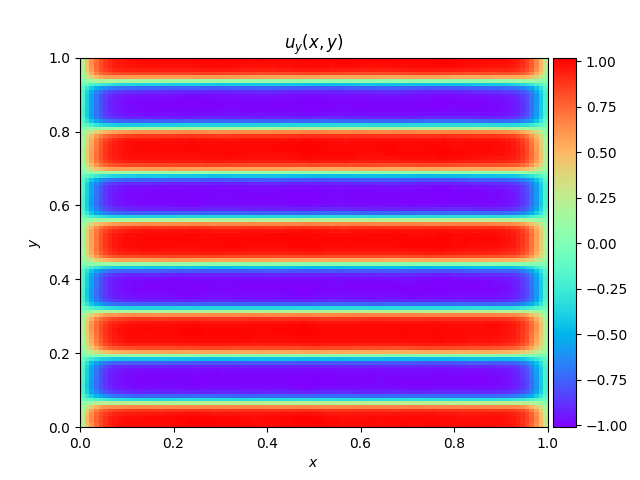}\label{subfig:uy_FF=3_eps=0.1by4}}
	\subfigure[$i = 4$]{\includegraphics[width=0.31\textwidth]{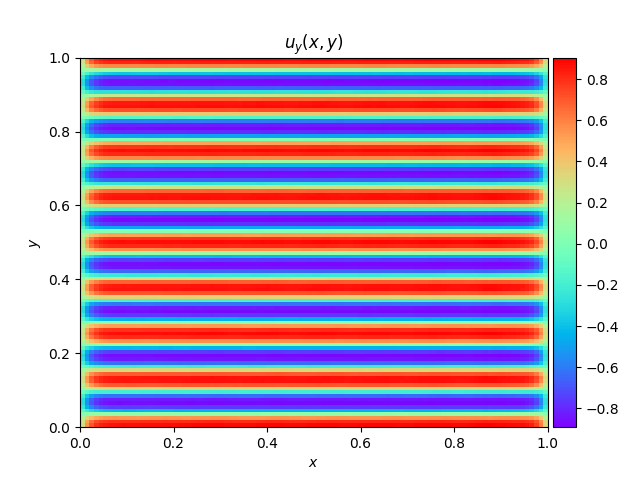}\label{subfig:uy_FF=4_eps=0.1by4}}
	\caption{DRM approximation to~\eqref{eqn:2D_Problem} with $3 \times 128$ NN and Fourier feature of frequency $\delta({\bf x}) = \left[{\bf x}, \sin(2^i\pi {\bf x}),\cos(2^i\pi {\bf x})\right]$ with $\varepsilon = 0.1/4$, $\eta = 1.0\times 10^{-4}$ after $300000$ epochs.}
	\label{fig:2D_problem_FF_eps=0.1by4}
\end{figure}

When increasing $\varepsilon$ further, we observe that the DRM method generates minimizing sequences with smoother interfacial transitions and larger microstructure length scales as shown in Fig.~\ref{fig:2D_problem_FF_eps=0.1by4}.  Additionally, we observe that, for $\varepsilon = 0.1/4$, when applying Fourier mapping of frequency $4\pi$ and $8\pi$, the DRM generates the same sequence (with 8 transitions) while the same Fourier mappings  and different values of $\varepsilon$ ($\varepsilon = 0.1/16$ and $\varepsilon = 0$) generate sequences with $8$ and $15$ transitions respectively (see Figs.~\ref{subfig:uy_FF=3_epsbar=0.0}, ~\ref{subfig:uy_FF=3_eps=0.1by16} and~\ref{subfig:uy_FF=3_eps=0.1by4}). A similar behavior is observed when applying a Fourier mapping of frequency $16\pi$: DRM generates a sequence with $16$ twin bands when $\varepsilon = 0.1/4$ and a sequence with $32$ twin bands for smaller values of $\varepsilon$ (compare Figs.~\ref{subfig:uy_FF=4_epsbar=0.0} with Figs.~\ref{subfig:uy_FF=4_eps=0.1by16} and ~\ref{subfig:uy_FF=4_eps=0.1by4}). This is perhaps not surprising since the regularization term imposes an upper bound on the number of interfaces that can be generated for a value of $\varepsilon$. 

\section{Conclusions} 
\label{sec:Concl}
This work employs DRM in conjunction with Fourier feature mapping (DRM\&FM) to solve non-convex minimization problems relevant in microstructure applications. We consider three benchmark problems: two minimization problems in $1D$ given by~\eqref{eqn:1D_Problem_1} and~\eqref{eqn:1D_Problem_2} and one in $2D$ given by~\eqref{eqn:2D_Problem}. These problems are challenging to solve since they often do not possess a global minimum (see~\eqref{eqn:1D_Problem_2} \& \eqref{eqn:2D_Problem}) or a global minimum exists (as in~\eqref{eqn:1D_Problem_1}), but there exist multiple functions that can yield such minimum.   

To tackle these challenges, we employ DRM in conjunction with Fourier feature mapping to generate high frequency, multiscale solutions. The method uses a DNN comprised of an input layer, a Fourier feature mapping of the form $\delta({\bf x}) =\left[{\bf x}, \sin(2^i\pi {\bf x}),\cos(2^i\pi {\bf x})\right]$, multiple hidden layers and an output layer.  Utilizing NTK theory, we demonstrate that the DRM as implemented in~\cite{Chen&Rosakis2023} suffers from spectral bias pathology: the rate at which the DNN learns minimizing solutions is determined by the largest eigenvalue of the NTK where $\lambda_i\gg 0$. To explore multiple solutions effectively, a desirable NTK should have eigenvalues $\lambda_i \approx 0$ to avoid spectral bias pathology.

Our heuristic analysis shows that the application of Fourier feature mapping results in a quadratic decay NTK eigenspectrum $\lambda_i \approx 0$, enabling our method DRM\&FM to generate high frequency, multiscale solutions. Simulations confirm the effectiveness of DRM\&FM in generating such solutions for all three benchmark problems. In contrast to the method proposed in~\cite{Chen&Rosakis2023}, simply increasing the depth of the neural network does not produce high-frequency solutions for our benchmark problems. However, our approach achieves this by keeping the network depth fixed and incorporating a Fourier mapping. 

While minimizing solutions may appear noisy without a regularization term, this capability still represents a significant advantage over the Finite Element Method (FEM). However, solving these types of problems remains challenging due to the rough energy landscape, which lacks well-defined minima and can hinder the algorithm’s training and solution generation. To address this issue, we considered a regularized minimization problem in 2D. We observed that incorporating a regularization term (Sec.~\ref{sec:2D_problem_epsilon}) smooths the energy landscape, facilitates training and produces symmetric, smooth solutions for small values of $\varepsilon$. As the value of $\varepsilon$ increases, we observe that the solutions generated by the method are low-frequency solutions.  

While DRM with Fourier mapping presents a mesh-free and computationally efficient algorithm, its nonlinear nature lacks a theoretical foundation to quantify solution accuracy for the considered minimization problems. We encourage the research community to develop such a theory in the near future.

\section{Acknowledgment}
Research was sponsored by the Army Research Laboratory and was accomplished under
Cooperative Agreements Number W911NF-22-2-0090 and W911NF-23-2-0139. The views and conclusions contained in this
document are those of the authors and should not be interpreted as representing the official
policies, either expressed or implied, of the Army Research Laboratory or the U.S. Government.
The U.S. Government is authorized to reproduce and distribute reprints for Government purposes
notwithstanding any copyright notation herein.

\appendix
\section{Derivation of the gradient dynamics}\label{app:gradient-evolution}
We provide the detailed derivation of the gradient dynamics~\eqref{eqn:gradient-evolution}.
Recall the notations
\[
\bar{U}_n(\theta) = [\bar{u}_n(x_n; \theta), \bar{u}_n^{\prime}(x_n; \theta)]^{\top} \in \mathbb{R}^{2\times 1},
\]
\[
\bar{U}_{\mathcal{X}}(\theta) = [\bar{U}_1(\theta), \ldots, \bar{U}_{|\mathcal{X}|}(\theta)]^{\top} \in \mathbb{R}^{ 2|\mathcal{X}| \times 1},
\]
\[
\bar{W}_n(\theta) =
W(x, \bar{u}({\bf x}; \theta),  \bar{u}^{\prime}({\bf x}; \theta)) \in \mathbb{R},\]
\[
\bar{W}_{\mathcal{X}}(\theta) = [\bar{W}_1(\theta), \ldots, \bar{W}_{|\mathcal{X}|}(\theta)] \in \mathbb{R}^{|\mathcal{X}| \times 1},
\]
\[ \nabla_U \bar{W}_n(\theta) = [\partial_{\hat{u}} \bar{W}_n(\theta), \partial_{\hat{u}^{\prime}} \bar{W}_n(\theta)]^{\top}
\in 
\mathbb{R}^{2 \times 1},\]
\[
\nabla_U \bar{W}_{\mathcal{X}}(\theta) = [\nabla_U \bar{W}_1(\theta), \ldots, \nabla_U \bar{W}_{|\mathcal{X}|}(\theta)]^{\top} \in \mathbb{R}^{2|\mathcal{X}| \times 1}.
\]
A simple application of the chain rule leads to 
\begin{equation}
	\begin{split}
		\frac{d[\nabla_{U} \bar{W}_{\mathcal{X}}(\theta(t))]}{dt}
		=
		\begin{bmatrix}
			\frac{d\partial_u \bar{W}_1(\theta(t))}{dt} \\
			\frac{d\partial_{u^{\prime}} \bar{W}_1(\theta(t))}{dt} \\
			\vdots\\
			\frac{d\partial_u \bar{W}_{|\mathcal{X}|}(\theta(t))}{dt} \\
			\frac{d\partial_{u^{\prime}} \bar{W}_{|\mathcal{X}|}(\theta(t))}{dt} 
		\end{bmatrix}
		=
		\begin{bmatrix}
			\partial_{uu}^2 \bar{W}_1 \frac{d \bar{u}_1(\theta(t))}{dt} + \partial_{uu^{\prime}}^2 \bar{W}_1 \frac{d \bar{u}_1^{\prime}(\theta(t))}{dt}\\
			\partial_{u^{\prime}u}^2 \bar{W}_1 \frac{d \bar{u}_1(\theta(t))}{dt} + \partial_{u^{\prime}u^{\prime}}^2 \bar{W}_1 \frac{d \bar{u}_1^{\prime}(\theta(t))}{dt}\\
			\vdots\\
			\partial_{uu}^2 \bar{W}_{|\mathcal{X}|} \frac{d \bar{u}_{|\mathcal{X}|}(\theta(t))}{dt} + \partial_{uu^{\prime}}^2 \bar{W}_{|\mathcal{X}|} \frac{d \bar{u}_{|\mathcal{X}|}^{\prime}(\theta(t))}{dt}\\
			\partial_{u^{\prime}u}^2 \bar{W}_{|\mathcal{X}|} \frac{d \bar{u}_{|\mathcal{X}|}(\theta(t))}{dt} + \partial_{u^{\prime}u^{\prime}}^2 \bar{W}_{|\mathcal{X}|} \frac{d \bar{u}_{|\mathcal{X}|}^{\prime}(\theta(t))}{dt}
		\end{bmatrix} 
		=
		D_{\mathcal{X}}(\theta(t)) \frac{d\bar{U}_{\mathcal{X}}(\theta(t))}{dt},
	\end{split}
\end{equation}
where
\begin{equation*}
	D_{\mathcal{X}}(\theta(t)) = \text{diag}\left( \begin{bmatrix}
		\partial_{uu}^2 \bar{W}_n & \partial_{uu^{\prime}}^2 \bar{W}_n \\
		\partial_{u^{\prime}u}^2 \bar{W}_n & \partial_{u^{\prime}u^{\prime}}^2 \bar{W}_n
	\end{bmatrix}    \right)_{n=1, \ldots, |\mathcal{X}|} \in \mathbb{R}^{2|\mathcal{X}| \times 2|\mathcal{X}|}.
\end{equation*}
Further note that by following the same argument as for deriving~\eqref{eqn:loss-trajectory-NTK}, we have 
\begin{equation}
	\begin{split}
		\frac{d\bar{U}_{\mathcal{X}}(\theta(t))}{dt}
		=
		\nabla_{\theta} \bar{U}_{\mathcal{X}}(\theta(t)) \frac{d\theta(t)}{dt}
		=
		-\frac{\eta}{|\mathcal{X}|} M_{\mathcal{X}} \nabla_{U} \bar{W}_{\mathcal{X}}(\theta(t)).
	\end{split}
\end{equation}
We obtain the desired dynamics~\eqref{eqn:gradient-evolution} for
$\nabla_{U} \bar{W}_{\mathcal{X}}(\theta(t))$.

\section{The eigenspectrum of the Gram matrix}\label{app:eigenspectrum}
Let $K: D \times D \to \mathbb{R}$ be a symmetric positive definite kernel and 
define the Hilbert Schmidt integral operator 
\[
\mathcal{L}u(x) \triangleq \int_D K(x, x^{\prime}) u(x^{\prime}) \, dx.
\]
Given a dataset $\mathcal{X} = \{x_1, \ldots, x_n\} \subset D$ that is uniformly sampled over $D$, the Gram matrix induced by $K$, i.e., 
\[
M_{\mathcal{X}} = K(\mathcal{X}, \mathcal{X}),
\]
plays a central role in various kernel based regression tasks. 
Assuming $D$ is compact and $K$ is a Mercer kernel, the integral operator $\mathcal{L}$ admits a discrete spectrum and hence the following eigenvalue problem is well defined~\cite{williams2006gaussian}, 
\begin{equation}\label{eqn:eigen-L}
	\mathcal{L}u_k = \Lambda_k u_k, \qquad k = 1, 2, \ldots,
\end{equation}
where the eigenvalues $\Lambda_1 \geq \Lambda_2 \geq \ldots > 0$ and the eigenfunctions are orthonormal, i.e.,
\[
\int_D u_i(x) u_j(x) \, dx = \delta_{ij}.
\]
Evaluating~\eqref{eqn:eigen-L} at $\mathcal{X}$ leads to 
\begin{equation}\label{eqn:eigen-LX}
	\mathcal{L}{\bf u}_k = \Lambda_k {\bf u}_k, \qquad k = 1, 2, \ldots,
\end{equation}
where ${\bf u}_k = u_k(\mathcal{X}) \in \mathbb{R}^{n \times 1}$ and $\mathcal{L}{\bf u}_k = [\mathcal{L}u_k(x_1), \ldots, \mathcal{L}u_k(x_n)]^{\top}$.
Note that the integral operator $\mathcal{L}$ can be approximated by 
\[
\mathcal{L}u(x) \approx \mathcal{L}_nu(x) \triangleq \frac{1}{n} \sum_{i = 1}^n K(x, x_i)u(x_i)
\]
and hence we can approximately (for $n$ large) consider the 
eigenvalue problem 
\begin{equation}\label{eqn:eigen-Ln}
	\mathcal{L}_n u_k = \hat{\Lambda}_k u_k, \qquad k = 1, 2, \ldots, n,
\end{equation}
where $\hat{\Lambda}_k \approx \Lambda_k$ depends on the sample size $n$.
Evaluating the above equation at $\mathcal{X}$ leads to 
\begin{equation}\label{eqn:eigen-M}
	M_{\mathcal{X}} {\bf u}_k = n \hat{\Lambda}_k {\bf u}_k,\qquad k = 1, \ldots, n,
\end{equation}
where $\lambda_k \triangleq n \hat{\Lambda}_k$ is the $k$-th eigenvalue for the Gram matrix $M_{\mathcal{X}}$. 
Comparing~\eqref{eqn:eigen-LX} with~\eqref{eqn:eigen-M} leads to the connection between the eigenvalue of $\mathcal{L}$ and the eigenvalue of the Gram matrix $M_{\mathcal{X}}$,
\[
\Lambda_k = \lim_{n \to \infty} \frac{\lambda_k}{n}.
\]
Therefore, for large values of $n$, we have the approximation $\lambda_k \approx n \Lambda_k$ for $k = 1, \ldots, n$.


\bibliographystyle{elsarticle-num}
\bibliography{bibliography}

\begin{thebibliography}{10}
\expandafter\ifx\csname url\endcsname\relax
  \def\url#1{\texttt{#1}}\fi
\expandafter\ifx\csname urlprefix\endcsname\relax\def\urlprefix{URL }\fi
\expandafter\ifx\csname href\endcsname\relax
  \def\href#1#2{#2} \def\path#1{#1}\fi

\bibitem{bhattacharya2003microstructure}
K.~Bhattacharya, Microstructure of martensite: why it forms and how it gives
  rise to the shape-memory effect, Vol.~2, Oxford University Press, 2003.

\bibitem{dacorogna2007direct}
B.~Dacorogna, Direct methods in the calculus of variations, Vol.~78, Springer
  Science \& Business Media, 2007.

\bibitem{luskin1996computation}
M.~Luskin, On the computation of crystalline microstructure, Acta numerica 5
  (1996) 191--257.

\bibitem{Carstensen_2005}
C.~Carstensen, Ten remarks on nonconvex minimisation for phase transition
  simulations, Computer Methods in Applied Mechanics and Engineering 194~(2)
  (2005) 169--193.

\bibitem{gobbert1999discontinuous}
M.~K. Gobbert, A.~Prohl, A discontinuous finite element method for solving a
  multiwell problem, SIAM journal on numerical analysis 37~(1) (1999) 246--268.

\bibitem{carstensen2001numerical}
C.~Carstensen, Numerical analysis of microstructure, Theory and Numerics of
  Differential Equations: Durham 2000 (2001) 59--126.

\bibitem{bartels2004effective}
S.~Bartels, C.~Carstensen, K.~Hackl, U.~Hoppe, Effective relaxation for
  microstructure simulations: algorithms and applications, Computer Methods in
  Applied Mechanics and Engineering 193~(48-51) (2004) 5143--5175.

\bibitem{carstensen1997numerical}
C.~Carstensen, P.~Plech{\'a}{\v{c}}, Numerical solution of the scalar
  double-well problem allowing microstructure, Mathematics of Computation
  66~(219) (1997) 997--1026.

\bibitem{nicolaides1993computation}
R.~A. Nicolaides, N.~J. Walkington, Computation of microstructure utilizing
  young measure representations, Journal of intelligent material systems and
  structures 4~(4) (1993) 457--462.

\bibitem{aranda2001numerical}
E.~Aranda, P.~Pedregal, Numerical approximation of non-homogeneous, non-convex
  vector variational problems, Numerische Mathematik 89 (2001) 425--444.

\bibitem{carstensen2000numerical}
C.~Carstensen, T.~Roub{\'\i}{\v{c}}ek, Numerical approximation of young
  measuresin non-convex variational problems, Numerische Mathematik 84 (2000)
  395--415.

\bibitem{hornik1990universal}
K.~Hornik, M.~Stinchcombe, H.~White, Universal approximation of an unknown
  mapping and its derivatives using multilayer feedforward networks, Neural
  networks 3~(5) (1990) 551--560.

\bibitem{hornik1991approximation}
K.~Hornik, Approximation capabilities of multilayer feedforward networks,
  Neural networks 4~(2) (1991) 251--257.

\bibitem{raissi2019physics}
M.~Raissi, P.~Perdikaris, G.~E. Karniadakis, Physics-informed neural networks:
  A deep learning framework for solving forward and inverse problems involving
  nonlinear partial differential equations, Journal of Computational physics
  378 (2019) 686--707.

\bibitem{sirignano2018dgm}
J.~Sirignano, K.~Spiliopoulos, Dgm: A deep learning algorithm for solving
  partial differential equations, Journal of computational physics 375 (2018)
  1339--1364.

\bibitem{yu2018deep}
B.~Yu, et~al., The deep ritz method: a deep learning-based numerical algorithm
  for solving variational problems, Communications in Mathematics and
  Statistics 6~(1) (2018) 1--12.

\bibitem{han2017deep}
J.~Han, A.~Jentzen, et~al., Deep learning-based numerical methods for
  high-dimensional parabolic partial differential equations and backward
  stochastic differential equations, Communications in mathematics and
  statistics 5~(4) (2017) 349--380.

\bibitem{grohs2018proof}
P.~Grohs, F.~Hornung, A.~Jentzen, P.~Von~Wurstemberger, A proof that artificial
  neural networks overcome the curse of dimensionality in the numerical
  approximation of black-scholes partial differential equations, arXiv preprint
  arXiv:1809.02362 (2018).

\bibitem{rahaman2019spectral}
N.~Rahaman, A.~Baratin, D.~Arpit, F.~Draxler, M.~Lin, F.~Hamprecht, Y.~Bengio,
  A.~Courville, On the spectral bias of neural networks, in: International
  Conference on Machine Learning, PMLR, 2019, pp. 5301--5310.

\bibitem{wang2021eigenvector}
S.~Wang, H.~Wang, P.~Perdikaris, On the eigenvector bias of fourier feature
  networks: From regression to solving multi-scale pdes with physics-informed
  neural networks, Computer Methods in Applied Mechanics and Engineering 384
  (2021) 113938.

\bibitem{Chen&Rosakis2023}
X.~Chen, P.~Rosakis, Z.~Wu, Z.~Zhang, Solving nonconvex energy minimization
  problems in martensitic phase transitions with a mesh-free deep learning
  approach, Computer Methods in Applied Mechanics and Engineering 416 (2023)
  116384.

\bibitem{FourierFeatures2020}
M.~Tancik, P.~P. Srinivasan, B.~Mildenhall, S.~Fridovich-Keil, N.~Raghavan,
  U.~Singhal, R.~Ramamoorthi, J.~T. Barron, R.~Ng, Fourier features let
  networks learn high frequency functions in low dimensional domains, NeurIPS
  (2020).

\bibitem{geifman2020similarity}
A.~Geifman, A.~Yadav, Y.~Kasten, M.~Galun, D.~Jacobs, B.~Ronen, On the
  similarity between the laplace and neural tangent kernels, Advances in Neural
  Information Processing Systems 33 (2020) 1451--1461.

\bibitem{chen2020deep}
L.~Chen, S.~Xu, Deep neural tangent kernel and laplace kernel have the same
  rkhs, arXiv preprint arXiv:2009.10683 (2020).

\bibitem{Weinan}
E.~Weinan, B.~Yu, The deep ritz method: A deep learning-based numerical
  algorithm for solving variational problems, Communications in Mathematics and
  Statistics 6~(1) (2018) 1--12.

\bibitem{kingma2014adam}
D.~P. Kingma, J.~Ba, Adam: A method for stochastic optimization, arXiv preprint
  arXiv:1412.6980 (2014).

\bibitem{jacot2018neural}
A.~Jacot, F.~Gabriel, C.~Hongler, Neural tangent kernel: Convergence and
  generalization in neural networks, Advances in neural information processing
  systems 31 (2018).

\bibitem{du2018gradient}
S.~S. Du, X.~Zhai, B.~Poczos, A.~Singh, Gradient descent provably optimizes
  over-parameterized neural networks, arXiv preprint arXiv:1810.02054 (2018).

\bibitem{chizat2019lazy}
L.~Chizat, E.~Oyallon, F.~Bach, On lazy training in differentiable programming,
  Advances in neural information processing systems 32 (2019).

\bibitem{lee2019wide}
J.~Lee, L.~Xiao, S.~Schoenholz, Y.~Bahri, R.~Novak, J.~Sohl-Dickstein,
  J.~Pennington, Wide neural networks of any depth evolve as linear models
  under gradient descent, Advances in neural information processing systems 32
  (2019).

\bibitem{arora2019exact}
S.~Arora, S.~S. Du, W.~Hu, Z.~Li, R.~R. Salakhutdinov, R.~Wang, On exact
  computation with an infinitely wide neural net, Advances in neural
  information processing systems 32 (2019).

\bibitem{rahimi2007random}
A.~Rahimi, B.~Recht, Random features for large-scale kernel machines, Advances
  in neural information processing systems 20 (2007).

\bibitem{Muller1993}
S.~Muller, Singular perturbations as a selection criterion for periodic
  minimizing sequences, Calc. Var. Partial Differential Equations 1~(2) (1993)
  169--204.

\bibitem{Kohn&Otto}
R.~V. Kohn, F.~Otto, Small surface energy, coarse-graining, and selection of
  microstructure, Physica D: Nonlinear Phenomena 107~(2) (1997) 272--289.

\bibitem{Muller}
S.~Müller, Variational models for microstructure and phase transitions,
  Lecture Notes in Math. 1713 (1999) 85--210.

\bibitem{Kohn1992}
R.~V. Kohn, S.~M{\"u}ller, Relaxation and regularization of nonconvex
  variational problems, Rendiconti del Seminario Matematico e Fisico di Milano
  62~(1) (1992) 89--113.

\bibitem{Dondl2016}
P.~Dondl, B.~Heeren, M.~Rumpf, Optimization of the branching pattern in
  coherent phase transitions, Comptes Rendus Mathematique 354~(6) (2016)
  639--644.

\bibitem{williams2006gaussian}
C.~K. Williams, C.~E. Rasmussen, Gaussian processes for machine learning,
  Vol.~2, MIT press Cambridge, MA, 2006.

\end{thebibliography}


\end{document}